\documentclass[10pt,twocolumn,letterpaper]{article}

\usepackage{iccv}
\usepackage{times}
\usepackage{epsfig}
\usepackage{graphicx}
\usepackage{amsmath}
\usepackage{amssymb}
\usepackage{multirow}
\usepackage{dsfont}
\usepackage{cite}
\usepackage{subfigure}
\usepackage{color}
\usepackage{colortbl}
\usepackage{array}
\usepackage[breaklinks=true,bookmarks=false]{hyperref}
\iccvfinalcopy 

\def\iccvPaperID{2507} 

\begin{document}

\title{Future Localization from an Egocentric Depth Image}

\author{Hyu Soo Park~~~~~~~~~Yedong Niu~~~~~~~~~Jianbo Shi\\
University of Pennsylvania\\
{\tt\small \{hypar,yedniu,jshi\}@seas.upenn.edu}
}

\maketitle

\begin{abstract}
This paper presents a method for future localization: to predict a set of plausible trajectories of ego-motion given a depth image.   
We predict paths avoiding obstacles, between objects, even paths turning around a corner into space behind objects.
As a byproduct of the predicted trajectories of ego-motion, we discover in the image the empty space occluded by foreground objects. We use no image based features such as semantic labeling/segmentation or object detection/recognition for this algorithm. Inspired by proxemics, we represent the space around a person using an EgoSpace map, akin to an illustrated tourist map, that measures a likelihood of occlusion at the egocentric coordinate system. A future trajectory of ego-motion is modeled by a linear combination of compact trajectory bases
allowing us to constrain the predicted trajectory. We learn the relationship between the EgoSpace map and trajectory from the EgoMotion dataset providing in-situ measurements of the future trajectory. A cost function that takes into account partial occlusion due to foreground objects is minimized to predict a trajectory. This cost function generates a trajectory that passes through the occluded space, which allows us to discover the empty space behind the foreground objects. We quantitatively evaluate our method to show predictive validity and apply to various real world scenes including walking, shopping, and social interactions. 
\end{abstract}

\section{Introduction}

\begin{figure}[t]
  \centering  
    \includegraphics[width=0.48\textwidth]{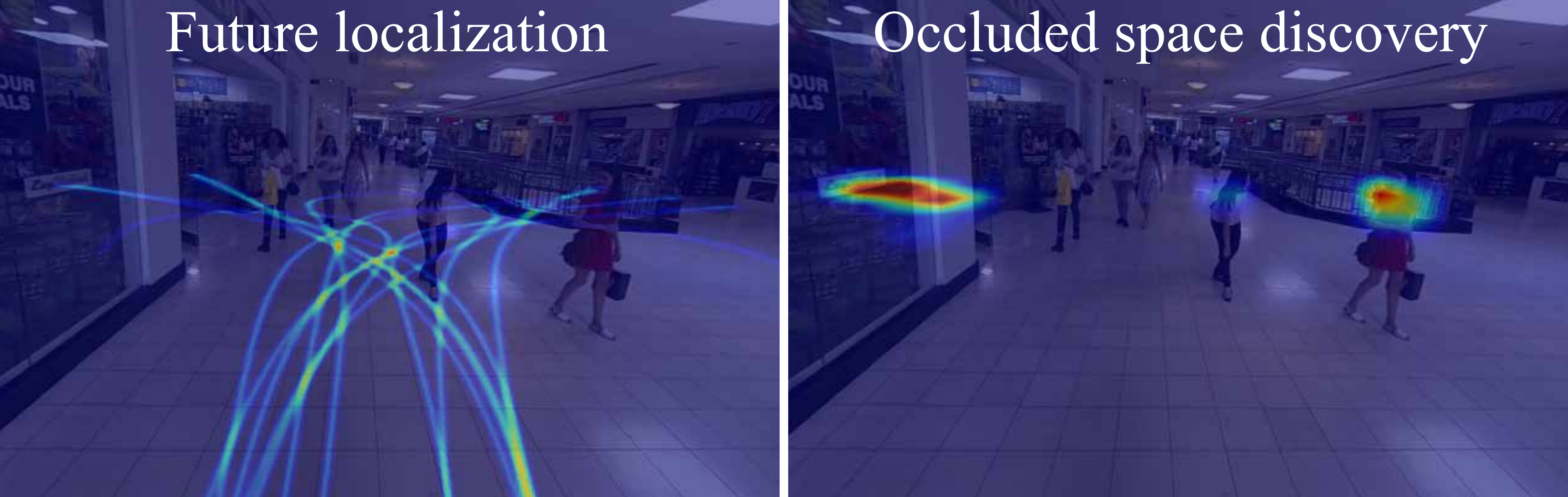}
  \caption{Where am I supposed to be after 5, 10, and 15 seconds? We present a method to predict a set of plausible trajectories given a first person depth image. As a byproduct of the predicted trajectories, the occluded space by foreground objects such as the space inside of the shop or behind the ladies are discovered.} 
  \label{Fig:teaser}
\end{figure}

Consider a dynamic scene such as Figure~\ref{Fig:teaser} where you, as the camera wearer, plan to pass through the corridor in the shopping mall while others walk in different directions. You need to plan your trajectory to avoid collisions with others and objects such as walls and fence. Looking ahead, you would plan a trajectory that enters into the shop by turning left at the corner although such space cannot be seen directly from your perspective.   

The fundamental problem we are interested in is {\em future localization}: where am I supposed to be after 5, 10, and 15 seconds?   This challenging task requires understanding of the scene in terms of a long term temporal human behaviors with respect to the spatial scene layout, with missing data due to occlusions.

We study the {\em future localization} problem using a first person depth (stereo) camera. We present a method to predict a set of plausible trajectories of ego-motion given a depth image captured from a egocentric view.   
As a byproduct of predicted trajectories, the occluded space behind foreground objects is discovered. 
Our method purely relies on the depth measurements, i.e., no image based features such as semantic labeling/segmentation or object detection/recognition are required. 

Inspired by proxemics~\cite{hall:1963}, we represent the space around a camera wearer using an EgoSpace map which reassembles an illustrated tourist map: an overhead map with objects seen from first person video projected onto it.   

\begin{figure*}[th]
  \centering  
      \subfigure[Ego-stereo cameras]{\label{Fig:stereo}\includegraphics[height=0.15\textheight]{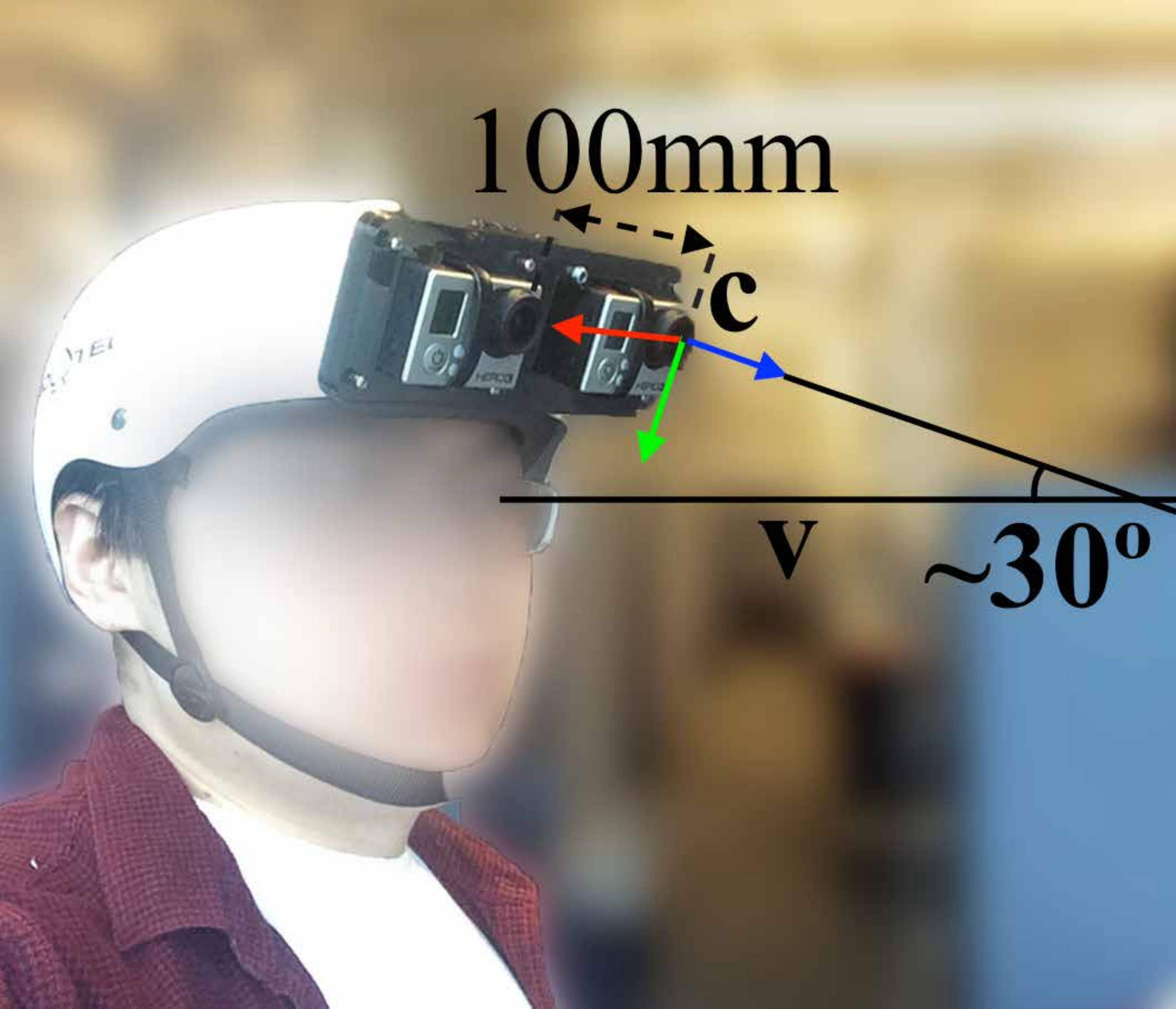}} 
      \subfigure[Geometry]{\label{Fig:notation}\includegraphics[height=0.15\textheight]{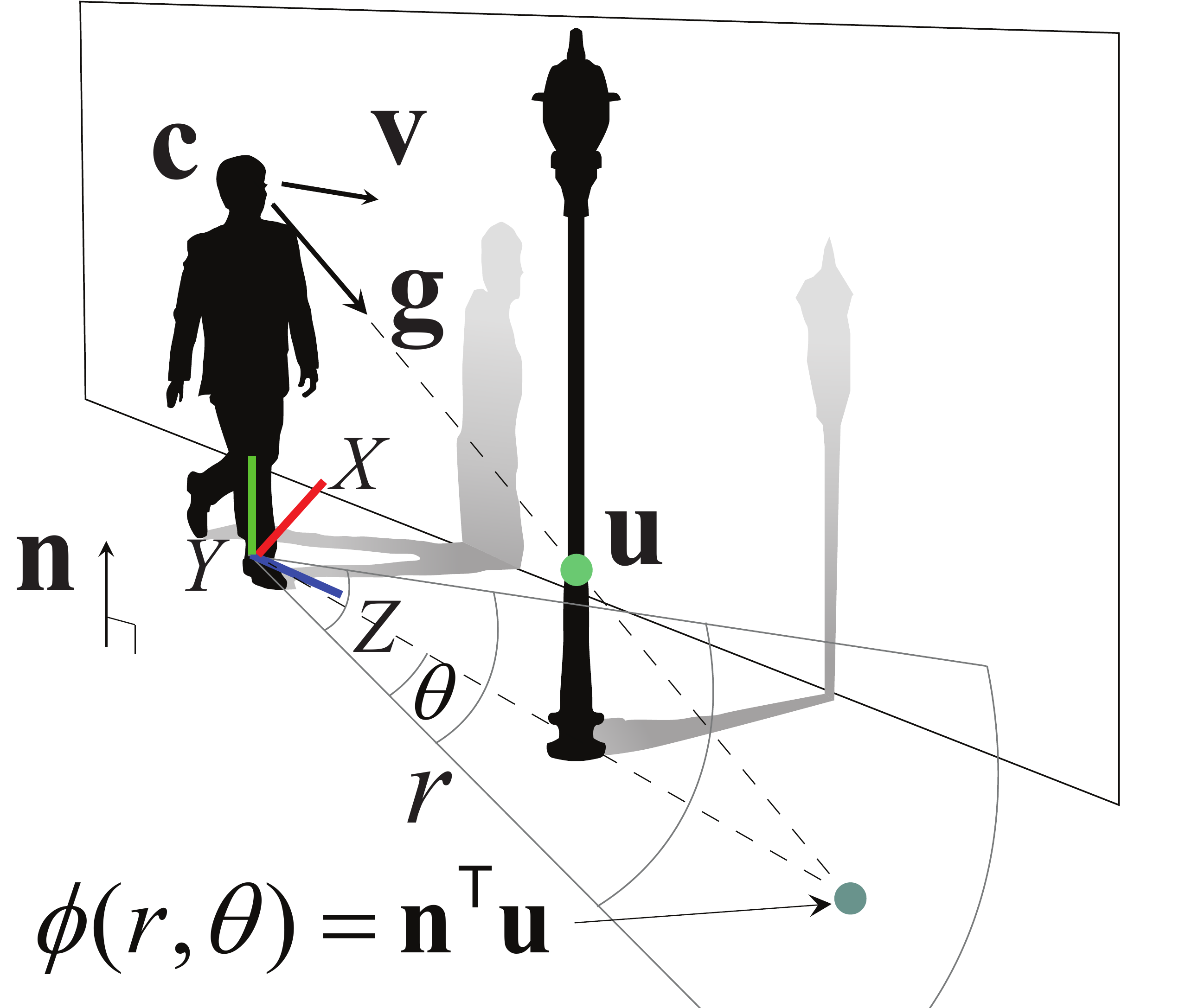}} 
      \subfigure[Depth image]{\label{Fig:depth}\includegraphics[height=0.15\textheight]{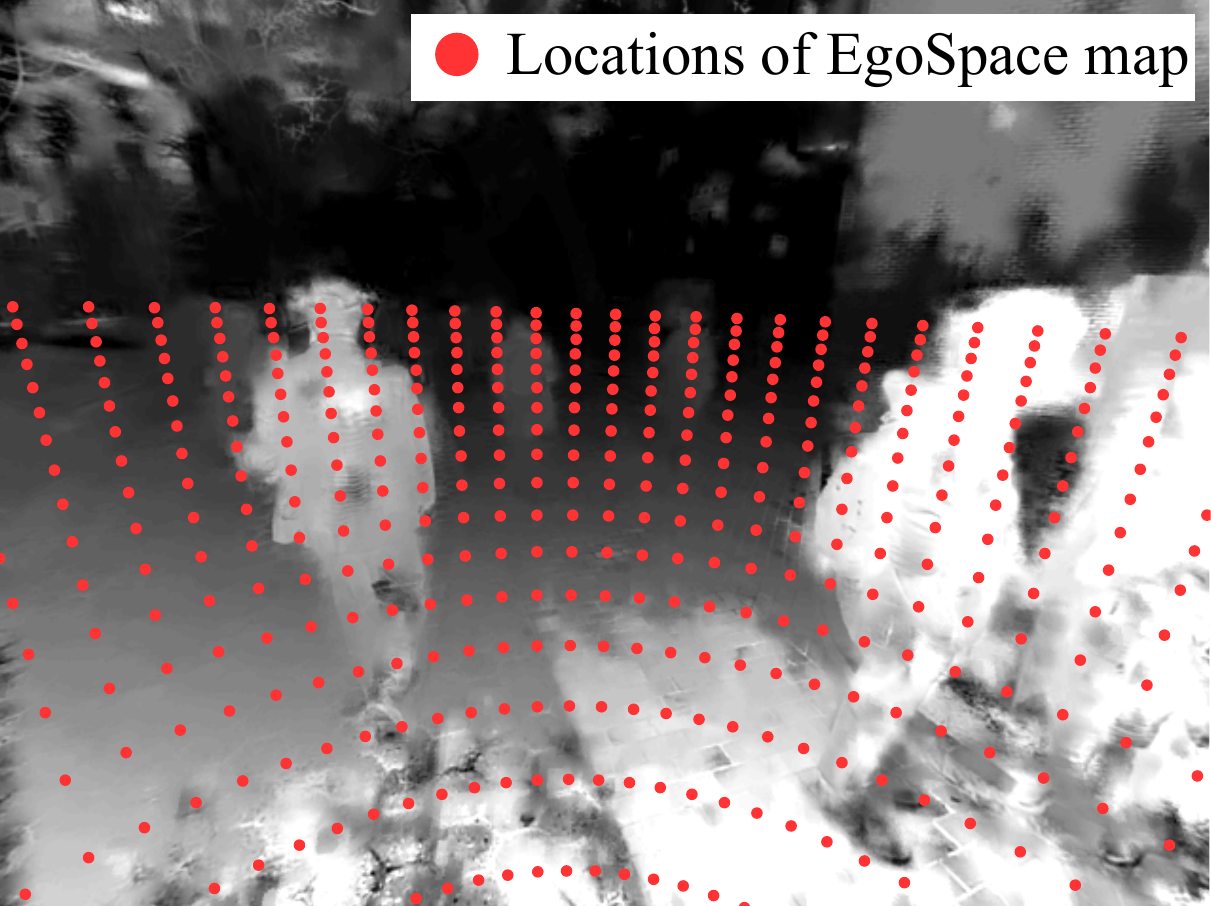}} 
            \subfigure[EgoSpace map]{\label{Fig:feature}\includegraphics[height=0.16\textheight]{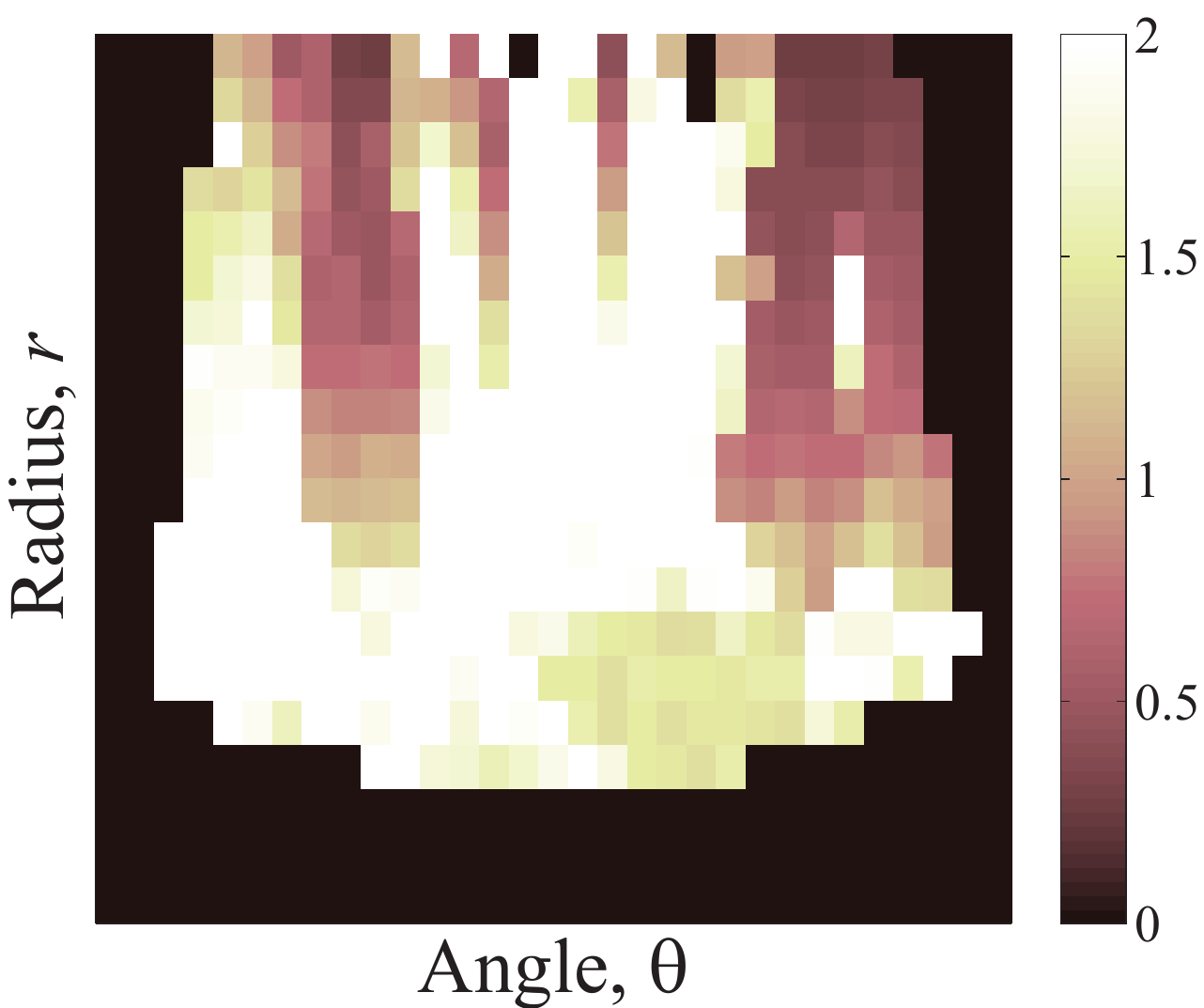}} 
  \caption{(a) We use ego-stereo cameras to capture our dataset where the depth image can be computed. Any depth sensor such as Kinect is complementary to our stereo setup. (b) Inspired by proxemics, we represent the space around a person using an EgoSpace map computed from (c) the depth image. (d) The EgoSpace map, $\phi(r,\theta)$, captures a likelihood of occlusion.} 
  \label{Fig:explanation}
\end{figure*}

A predictive {\em future localization} model, using the EgoSpace map, is learned from in-situ first person stereo videos from various life logging activities such as commutes, shopping, and social interactions. By leveraging structure from motion, camera trajectories are reconstructed.  These camera trajectories are associated with its depth image at each time instant, i.e., given the depth image, a future camera trajectory is precisely measured while the depth image is obtained by the stereo cameras\footnote{Any depth sensor such as Kinect and Creative Senz3D are complimentary to our depth measurement.} as shown in Figure~\ref{Fig:stereo}. 

In a training phase, we discriminatively learn the relationship between the EgoSpace map and future camera trajectory. We model a trajectory of ego-motion using a linear combination of compact trajectory bases. By the nature of the alignment between ego-motion and gaze direction, the trajectory is highly structured. We empirically show that 4$\sim$6 linear trajectory bases are sufficient enough to express all plausible trajectories of ego-motion with high precision (99$\%$ accuracy). This compact representation allows us to efficiently find a set of trajectories that are compatible with the associated depth image using EgoSpace map matching.  This provides an initialization of the predicted trajectories.  However, not all these `re-imagined' trajectories avoid objects in the current first person view.
We refine it by minimizing a cost function that takes into account compatibility between the obstacles in EgoSpace map and trajectory. This cost function explicitly models partial occlusion of a trajectory which allows us to discover the space behind foreground objects.

\noindent\textbf{Why EgoSpace map?} Two cues are strongly related to predict a trajectory of ego-motion, e.g., where is he or she going? (1) ego-cue: a vanishing point is often aligned with gaze direction; and 2D visual layout of the obstacles in the first person view implicitly encodes the semantics of the scene. 
(2) exo-cue: objects in a 3D scene such as road, buildings, and tables constrain the space where the wearer can navigate. Such cues can be explicitly extracted by an ego-depth image where the gaze direction of the wearer can be calibrated with respect to a ground plane (exocentric coordinate) while the depth provides obstacles with respect to the wearer (egocentric coordinate). Our EgoSpace map representation exploits these two cues where we measure depth from an egocentric view, and create an illustrated tourist map representation capturing both 2D visual arrangement of the obstacles (in first person view) and their 3D layout (in overhead view). This representation allows us to analyze and understand different scene types and gaze directions in the same coordinate system. 


\noindent\textbf{Contributions} To our best knowledge, this is the first paper that predicts ego-motion from a depth image without semantic scene labels or object detection via in-situ first person measurements. Core technical contributions of our paper are (a) a predictive model that describes a spatial distribution of objects with respect to an egocentric view, allowing us to register different scenes in a unified coordinate system; (b) a compact subspace representation of the predicted trajectories enabling a search for trajectory parameters feasible without explicit modeling of dynamics of human behaviors; (c) occluded space discovery through trajectory prediction; and (d) the EgoMotion dataset with a depth and its long term camera trajectory, which includes diverse daily activities across camera wearers. We evaluate our algorithm to predict ego-motion in real world scenes.

\section{Related Work}
Our framework lies an intersection between behavior prediction and egocentric vision.  
\subsection{Human Behavior Prediction}
Predicting where-to-go is a long standing task in behavioral science. This task requires to understand the interactions of agents with objects in a scene that afford a space to move. There is a large body of literature on human behaviors prediction algorithms. Pentland and Lin~\cite{pentland:1995} modeled human behaviors using a hidden Markov dynamic model to recognize driving patterns. Such Markovian model is an attractive choice to encode human behaviors because it reflects the way humans make a decision~\cite{ziebart:2008, lee:2011,levine:2012}. These models, especially partially observable Markov decision process (POMDP), have influenced motion planning in robotics~\cite{pineau:2007, kurniawati:2009, ragi:2013}. 

In computer vision, Ali and Shah~\cite{ali:2008} developed a flow field model that predicts spatial crowd behaviors for tracking extremely cluttered crowd scenes. Inspired by the social force model~\cite{helbing:1995}, Mehran et al.~\cite{mehran:2009} predicted pedestrian behaviors in a crowd scene to detect abnormal behaviors, and Pellegrini et al.~\cite{pellegrini:2009} used a modified model to track multiple agents. Ryoo~\cite{ryoo:2011} presented a bag-of-word approach to recognize social activities at the early stage of videos. Vu et al.~\cite{vu:2014} predicted plausible activities from a static scene by associating the scene statistics and labeled actions. In terms of the trajectory prediction task, our work is closely related with three path planning frameworks by Gong et al.~\cite{gong:2011}, Kitani et al.~\cite{kitani:2012}, and Alahi et al.~\cite{alahi:2014}. Gong et al. presented a method to generate multiple plausible trajectories of each agent in the scene constructed by homotopy classes, which allows them to produce a long term trajectory for visual tracking in crowd scenes.
Kitani et al. leveraged inverse optimal control theory to learn human preference with respect to the scene semantic labels, which enables them to predict the paths an agent follows. 
Alahi et al. introduced a geometric feature, social affinity model that captures a spatial relationship of neighboring agents to predict destinations of a crowd. 

Unlike previous methods that use semantic labels/segmentation or object detection/tracking which are often noisy in real world scenes, our measurements are a single depth image that can be reliably obtained by stereo cameras or depth sensors.  Estimating optimal parameters for Markovian models is often intractable. In contrast, our trajectory representation in a egocentric view can be encoded using compact trajectory bases, thus it makes learning tractable because of the reduced number of parameters.  

\subsection{Egocentric Vision}
A first person camera is an ideal camera placement to observe human activities because it reflects the attention of the camera wearer. This characteristics provides a powerful cue to understand human behaviors~\cite{fathi:2011,kitani:2011,fathi:2013,pirsiavash:2012,ryoo:2013}. 

Kitani et al.~\cite{kitani:2011} used scene statistics produced by camera ego-motion to recognize sport activities from a firse person camera. Traditional vision frameworks such as object detection, recognition, and segmentation frameworks are successfully integrated in first person data: Pirsiavash and Ramanan~\cite{pirsiavash:2012} recognized daily activities using deformable part models, Lee et al.~\cite{lee:2012} found important persons and objects, Fathi et al.~\cite{fathi:2011} discovered objects, and Li et al.~\cite{li:2013, li_cvpr:2013} segmented pixels corresponding to hands.
In a social setting, Fathi et al.~\cite{fathi:2012} presented a method to recognize social interactions by detecting gaze directions of people and Park et al.~\cite{park:2012} introduced an algorithm to reconstruct joint attention in 3D by leveraging 3D reconstruction of camera ego-motion. This reconstruction allows prediction of joint attention possible by learning the spatial relationship between a social formation and joint attention~\cite{park:2015}. 

Such characteristics of first person cameras were used to generate interesting applications in vision, graphics, and robotics. Lee et al.~\cite{lee:2012} summarized a life logging video, Xiong et al.~\cite{xiong:2014} detected iconic images using a web image prior. Arev et al.~\cite{arev:2014} used 3D joint attention to edit social video footages and Kopf et al.~\cite{kopf:2014} used 3D camera motion to generate a hyperlapse first person video. In robotics, Ryoo et al.~\cite{ryoo:2015} predicted human activities for human-robot interactions.

Unlike most previous methods, our task primarily focuses on predicting future behaviors by leveraging in-situ measurements from 3D reconstruction of camera ego-motion. This also allows us to tackle a more challenging problem---to discover an empty space that is not observable because of visual occlusion.

\section{Representation} \label{Sec:representation}

Inspired by proxemics~\cite{hall:1963}, we present a characterization of space with respect to the egocentric coordinate system, called EgoSpace map.  


\subsection{EgoSpace Map} \label{Sec:feaure}
EgoSpace Map is a representation for space experienced from first-person view but visualized in an overhead bird-eye map, akin to an illustrated tourist map.
 
It has three key ingredients.  First, we define an egocentric coordinate system centered at the feet location, the projection of the center of eyes onto the ground plane as shown in Figure~\ref{Fig:notation}. The normal direction, $\mathbf{n}$, of the ground plane is aligned with the $Y$-axis, and the height of the eye location is $h$, i.e., $\mathbf{c} = \left[\begin{array}{ccc}0&h&0\end{array}\right]^\mathsf{T}$ where $\mathbf{c}$ is the 3D location of the center of eyes. 
The gaze direction defines tangential directions of the ground plane: the $Z$-axis is aligned with the projection of the gaze direction, $\mathbf{v}$, i.e., $\mathbf{v} = \left[\begin{array}{ccc}0&v_y&v_z\end{array}\right]^\mathsf{T}$.



Second, the EgoSpace encodes depth cue from a first person view onto an overhead view on the ground plane.  Using a log-polar $\phi(r,\theta)$ parametrization of the X-Z (ground) plane, we define EgoSpace Map as a function  $\phi:\mathds{R}\times\mathds{S}^1\rightarrow\mathds{R}$, measuring likelihood of occlusion introduced by foreground objects from the gaze direction. One can think of the eye gaze is a light source shining on foreground objects casting shadows onto the ground plane.  On the shadow image we record the object height which is proportional to the occlusion likelihood.

Formally, $\phi(r,\theta)$ measures the height of the point, $\mathbf{u}$, from the ground plane that intersects the ray, $\mathbf{g}$, from the center of eyes, $\mathbf{c}$, to $(r, \theta)$ with an occluding object, $\mathcal{O}$, i.e., 
\begin{eqnarray}
\phi(r,\theta) &=& \mathbf{u}^\mathsf{T}\mathbf{n},
\end{eqnarray}
where $\mathbf{u} = \min_{\lambda \in \mathcal{L}} \lambda\mathbf{g}+\mathbf{c}$ such that $\mathcal{L} = \{\lambda | \lambda\mathbf{g}+\mathbf{c} \subset \cup_{i=1}^I \mathcal{O}_i, \lambda > 0\}$. $\{\mathcal{O}_i\}_{i=1}^O$ is a set of objects in the scene. 


We discretize the polar coordinate system by uniform sampling in angle between $\pi/6$ and $5\pi/6$ and uniform sampling in the inverse of radius which results in uniform sampling in the egocentric view as shown in Figure~\ref{Fig:depth}. Note that the locations to measure the EgoSpace map are almost radially uniform from the first person view point. 
 Figure~\ref{Fig:feature} shows the EgoSpace map for Figure~\ref{Fig:depth}. 

For {\em future localization}, ground plane provides a free space for us to move into. On the EgoSpace map, $\phi(r,\theta)=0$ if from the first person view the point $(r,\theta)$ lies on the ground plane.
More interestingly, the space {\em behind} an object also indicates  potential places to navigate.  Since the EgoSpace map is represented in the ground plane, not in first person view, the space behind the object are marked as occluded area (the right few columns of the map).


Third, the area outside of a first person view depth image boundary is set to $\phi_{\rm max}=2$m.  On the EgoSpace map, shape of the mask is uniquely defined by the gaze direction (roll and pitch angles of the head direction).
For example, Figure~\ref{Fig:depth} shows a case where the wearer is looking ahead almost parallel to the ground, the ground area close to the wearer ($r < 0.5$m) was not visible
e.g., $\phi(r < 0.5{\rm m},\theta)$ is marked as $\phi=\phi_{\rm max}$.  If the wearer is looking down, the masked area on EgoSpace would be for large values of $r$.

The EgoSpace representation supports learning {\em future localization} from first person videos by combining cues from 3D scene geometry and gaze direction.  Its benefits include: 1) the gaze direction normalized coordinate system provides a common 3D reference frame to learn; 2) overhead view representation removes the variations in first person 3D experience due to the head's pitch angle, 3) the log-polar encoding and sampling gives more importance to nearby space, and 4) the depth masking encodes implicitly both roll and pitch angle of head, making it more situation aware.


\subsection{Compact Trajectory Representation}

Let $\mathbf{X}=\left[\begin{array}{ccccc}x_1&z_1&\cdots&x_F&z_F\end{array}\right]^\mathsf{T} \in \mathds{R}^{2F}$ be a 2D trajectory on the  
ground plane of the egocentric coordinate system, where $F$ is the number of future frames to predict and $x_i$ and $z_i$ are two coordinates at the $i^{\rm th}$ time instance as shown in Figure~\ref{Fig:notation}. In practice, this trajectory can be obtained by projecting 3D camera poses between the $f+1$ and $f+F$ time instances at the $f^{\rm th}$ time instant onto the ground plane. This allows us to represent all trajectories in the same egocentric coordinate system, which are normalized by gaze direction because the $Z$ axis is aligned with the gaze direction. 

\begin{figure}[th]
  \centering  
      \subfigure[Ego-motion trajectories]{\label{Fig:traj}\includegraphics[height=0.145\textheight]{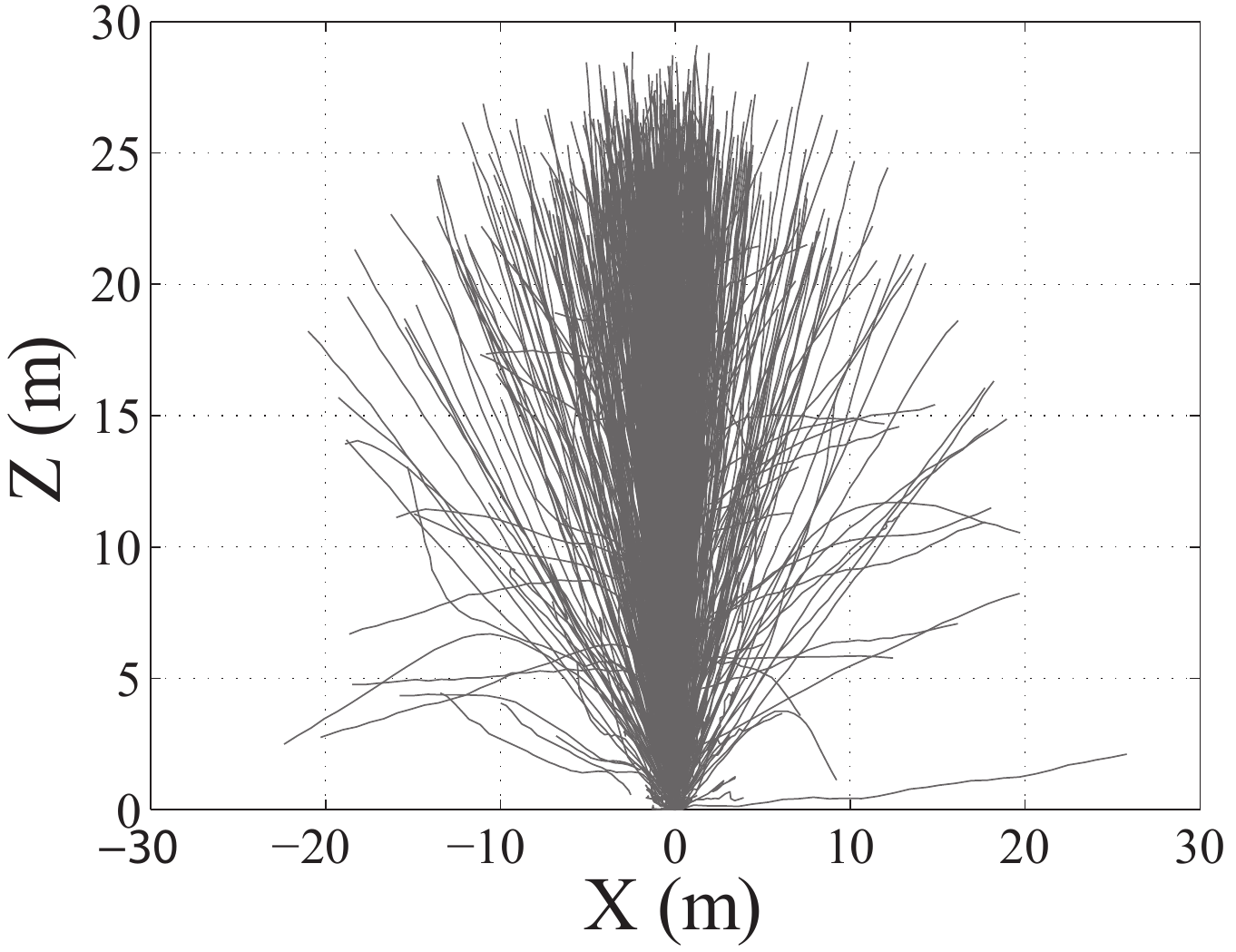}} 
      \subfigure[Reconstruction error]{\label{Fig:reconstruction_error}\includegraphics[height=0.145\textheight]{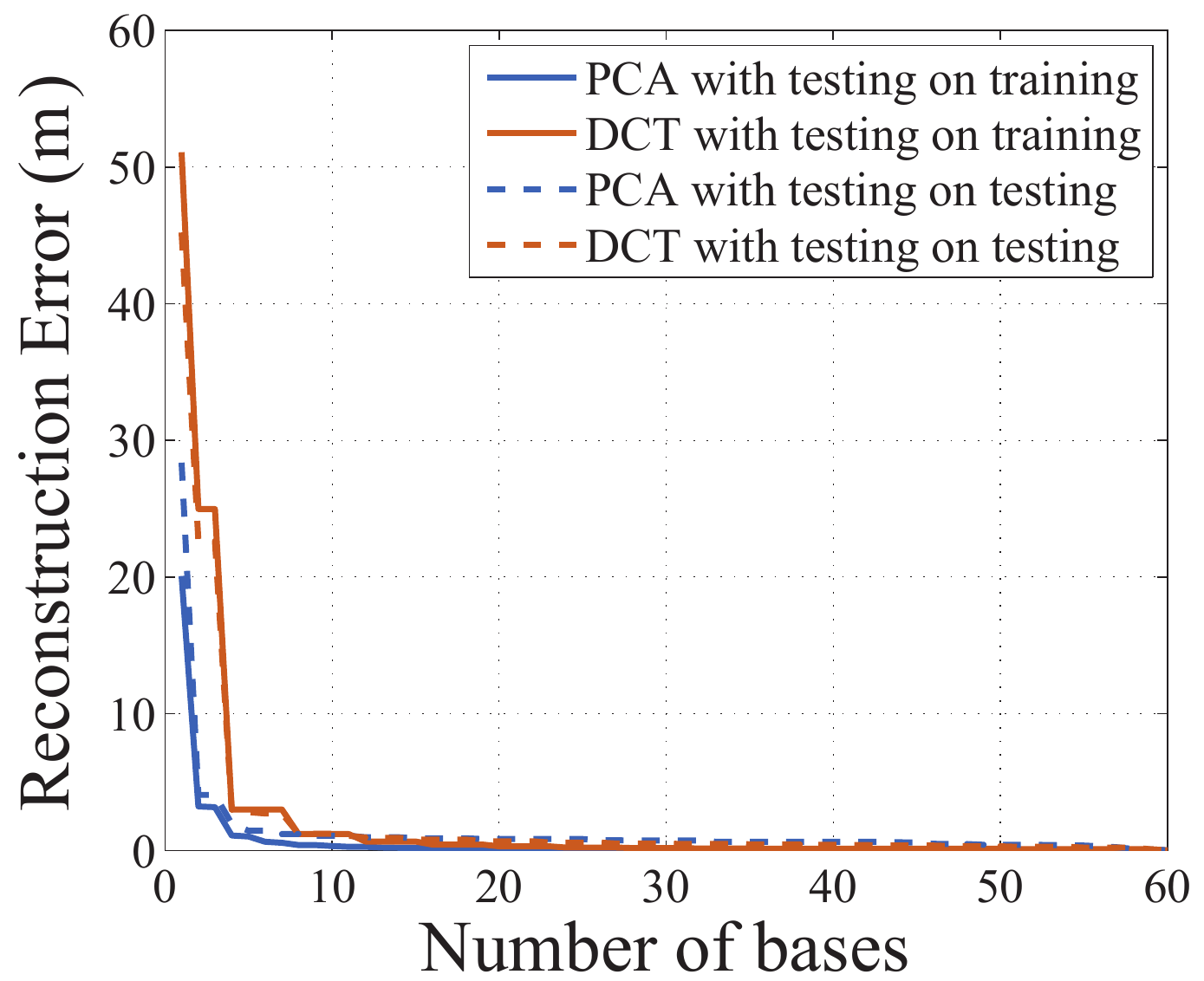}} 
  \caption{(a) We register all trajectories in an ego-centric coordinate system, which results in highly redundant trajectories that can be represented by a linear combination of (b) compact trajectory bases.} 
  \label{Fig:explanation}
\end{figure}

The gaze direction normalized trajectory is highly compressible.  Most trajectory of ego-motion can be encoded using a linear combination of trajectory bases learned using Principal Coordinate Analysis (PCA) from the EgoMotion dataset described in Section~\ref{Sec:data}:
\begin{eqnarray}
\mathbf{X} &=& \mathbf{B}\boldsymbol{\beta} + \overline{\mathbf{X}},
\end{eqnarray}
where $\overline{\mathbf{X}}$ is a mean trajectory and $\mathbf{B} \in \mathds{R}^{2F\times K}$ is a collection of trajectory bases, i.e., each column of $\mathbf{B}$ is a trajectory basis where $K$ is the number of basis. In practice, $K$ is selected as 4$\sim$6 which can express all ego-motion trajectories with $99\%$ accuracy as shown in Figure~\ref{Fig:traj} and Figure~\ref{Fig:reconstruction_error}.  $\boldsymbol{\beta} \in \mathds{R}^{K}$ is the trajectory coefficient, which is the low dimensional parametrization of $\mathbf{X}$. In Figure~\ref{Fig:reconstruction_error}, we compare reconstruction error produced by PCA bases and DCT generic bases~\cite{akhter:2011}. 

\section{Prediction}
A trajectory of ego-motion is associated with an EgoSpace map, i.e., given a depth image, we know how we explored the space in the training data (Section~\ref{Sec:data}). By leveraging a computational representation of egocentric space and trajectory described in Section~\ref{Sec:representation}, in this section, we present a method to predict a set of plausible trajectories given an EgoSpace map and to discover the occluded space using the predicted trajectories. 

\subsection{Ego-motion Prediction}
Estimating $\mathbf{X}$ that conforms to a depth image is to find a path that stays in the ground plane minimizing the following cost function along the trajectory:
\begin{eqnarray}
&\underset{\boldsymbol{\beta}}{\operatorname{minimize}}~~ \sum_i^{F} \widetilde\phi\left(\mathbf{B}_i \boldsymbol{\beta}\right), \label{Eq:min1}
\end{eqnarray}
where $\widetilde\phi:\mathds{R}^2\rightarrow\mathds{R}$ is the Cartesian coordinate representation of the EgoSpace map, $\phi$ and $\mathbf{B}_i \in \mathds{R}^{2\times K}$ is a matrix composed of the $(2(i-1)+1)^{\rm th}$ and $2i^{\rm th}$ rows of $\mathbf{B}$. Therefore, $\mathbf{B}_i\boldsymbol{\beta}$ is the point $(x_i, z_i)$ at the $i^{\rm th}$ time instant.

Equation~(\ref{Eq:min1}) finds a trajectory that stays on the ground given a depth image. This approach has been used in robotics communities for various path planning tasks. However, this does not take into account the trajectory that is partially occluded by objects because the occluded part of the trajectory always produces higher cost. Instead, we introduce a novel cost function that minimizes a trajectory cost difference between the given depth image and the retrieved depth image from the database:
\begin{eqnarray}
&\underset{\boldsymbol{\beta}}{\operatorname{minimize}}~~ \sum_i^{F} \max\left(0, \widetilde\phi\left(\mathbf{B}_i\boldsymbol{\beta}\right) - \widetilde\phi_D\left(\mathbf{B}_i\boldsymbol{\beta}_D\right)\right), \label{Eq:min}
\end{eqnarray}
where $\widetilde\phi_D$ and $\boldsymbol{\beta}_D$ are the EgoSpace map and trajectory parameter retrieved from the training dataset. This minimization finds a partially occluded trajectory as long as there exists a trajectory in the database that has similar occlusion cost. 

There exist infinite number of trajectories that are compatible with a given EgoSpace map. More importantly, the cost function in Equation~(\ref{Eq:min}) is nonlinear where an initialization of the solution is critical. 

We initialize $\boldsymbol{\beta}$ using a trajectory retrieved from the training data by EgoSpace map matching. The dataset is divided into 3 gaze directions (3 pitch angles) to reduce the false matches dominated by the area beyond the depth image. Given an EgoSpace map, k-nearest neighbors (KNN) are found using K-d tree~\cite{muja:2014}. Other search or planning methods such as structured SVM~\cite{tsochantaridis:2005} and Rapidly Exploring Random Tree (RRT)~\cite{lavalle:1998} can be complimentary to the KNN search. 

\subsection{Occluded Space Discovery}
The predicted trajectories of ego-motion allow us to discover the hidden space occluded by foreground objects because the trajectories can be still predicted in the hidden space. We build a likelihood map of the occluded space as follows:
{\small
\begin{eqnarray}
\psi(\mathbf{x}) = \frac{\sum_{j=1}^J\sum_{i=1}^F \exp\left(-\|\mathbf{x}-\mathbf{B}_i\boldsymbol{\beta}_j\|^2/2\sigma^2\right) \widetilde\phi\left(\mathbf{B}_i\boldsymbol{\beta}_j\right)}{\sum_{j=1}^J \sum_{i=1}^F\exp\left(-\|\mathbf{x}-\mathbf{B}_i\boldsymbol{\beta}_j\|^2/2\sigma^2\right)}, \label{Eq:discovery}
\end{eqnarray}}
where $\psi(\mathbf{x})$ is the likelihood of the occluded space that a trajectory can pass through at the evaluating point $\mathbf{x} \in \mathds{R}^2$ in the ground. $\boldsymbol{\beta}_j$ is the $j^{\rm th}$ predicted trajectories, $J$ is the number of predicted trajectories, and $\sigma$ is the bandwidth for the Guassian kernel. Equation~(\ref{Eq:discovery}) takes into account the likelihood of the predicted trajectories weighted by the likelihood of the occlusion. $\psi(\mathbf{x})$ is high when many trajectories are predicted at $\mathbf{x}$ while $\widetilde\phi(\mathbf{x})$ is high.




\section{EgoMotion Dataset} \label{Sec:data}
We present a new dataset, EgoMotion dataset, captured by first person stereo cameras. This dataset includes various indoor and outdoor scenes such as Park, Malls, and Campus with various activities such as walking, shopping, and social interactions.

\subsection{Data Collection}

A stereo pair of GoPro Hero 3 (Black Edition) cameras with 100mm baseline are used to capture EgoMotion dataset as shown in Figure~\ref{Fig:stereo}. All videos are recorded at 1280$\times$960 with 100fps. The stereo cameras are calibrated prior to the data collection and synchronized manually with a synchronization token at the beginning of each sequence.

\noindent\textbf{Depth Computation} We compute disparity between the stereo pair after stereo rectification. A cost space of stereo matching is generated for each scan line and match each pixel by exploiting dynamic programming in a coarse-to-fine manner.



\noindent\textbf{3D Reconstruction of Ego-motion}
We reconstruct a camera trajectory using a standard structure from motion pipeline with a few modifications to handle a large number of images\footnote{A 30 minute walking sequence at a 30 fps reconstruction rate produces HD 108,000 images.}. We partition the dataset such that each dataset includes less than 500 images with sufficient overlap with neighbor image sets (100 image overlap). We reconstruct each dataset independently and merge them by minimizing cross reprojection error between two dataset, i.e., a point in one dataset is reprojected to a camera in the other dataset. Then, we project the reconstructed camera trajectory onto the ground plane estimated by fitting a plane using RANSAC~\cite{fischler:1981}.

\noindent\textbf{Scenes}
We collect both indoor and outdoor data, which consists of 21 scenes with 55,933 frames of 7.7 hours long in total, including walking on campus, in parks and downtown streets, shopping in the mall, cafe and grocery, as well as taking public transportation. The data consists of various activities (walking, talking, and shopping), scenes (campus, park, malls, and downtown streets), cities, and time. We also collect repeated daily routines multiple times at a campus. The dataset is summarized in Table~\ref{table:dataset}.

\begin{table*}[ht]
\scriptsize
\centering
\begin{tabular}{>{\centering\arraybackslash}m{1.0cm}| *{8}{>{\centering\arraybackslash}m{1.5cm}} >{\centering\arraybackslash}m{0.1cm}}
\hline

Image Disparity &\centering\includegraphics[width=0.9cm]{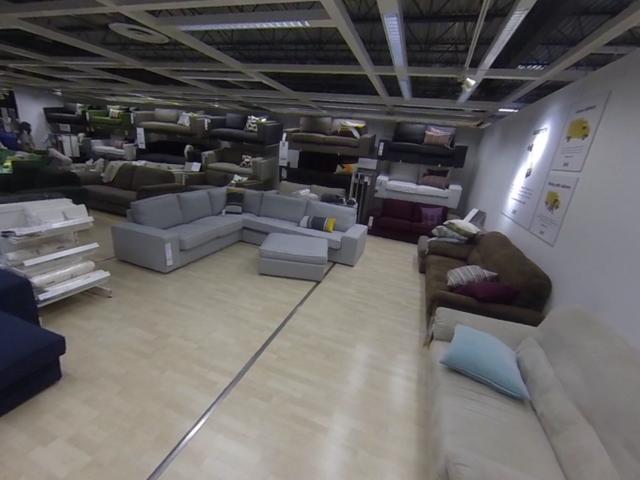}\includegraphics[width=0.9cm]{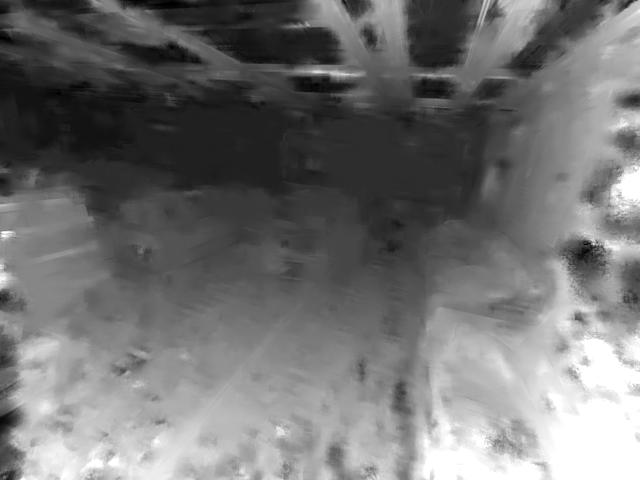} & \centering\includegraphics[width=0.9cm]{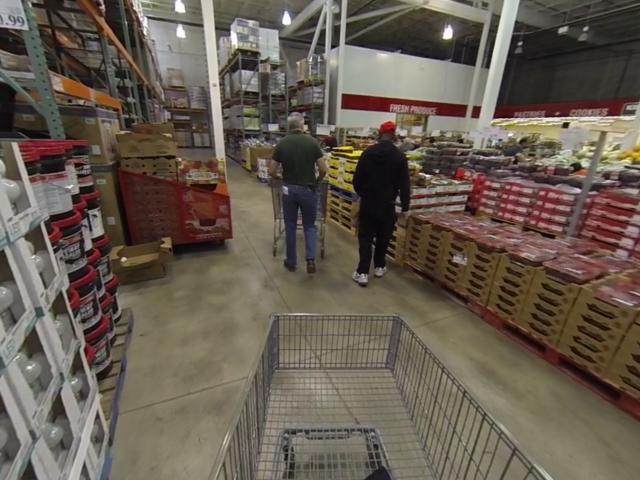}\includegraphics[width=0.9cm]{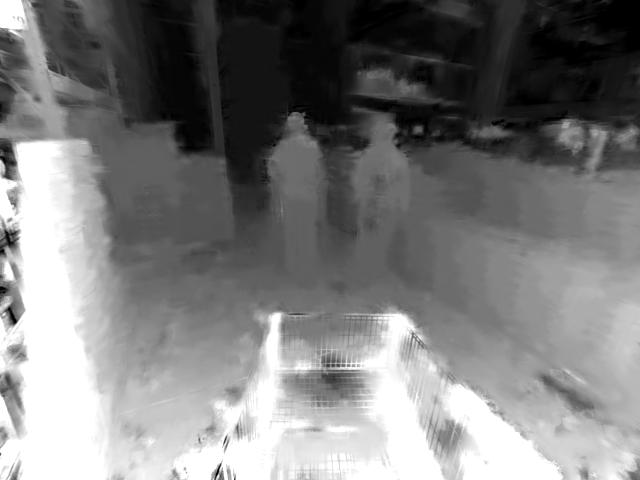} & \centering\includegraphics[width=0.9cm]{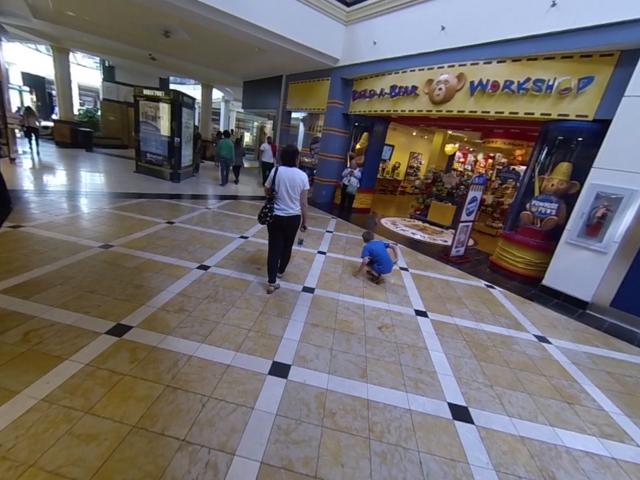}\includegraphics[width=0.9cm]{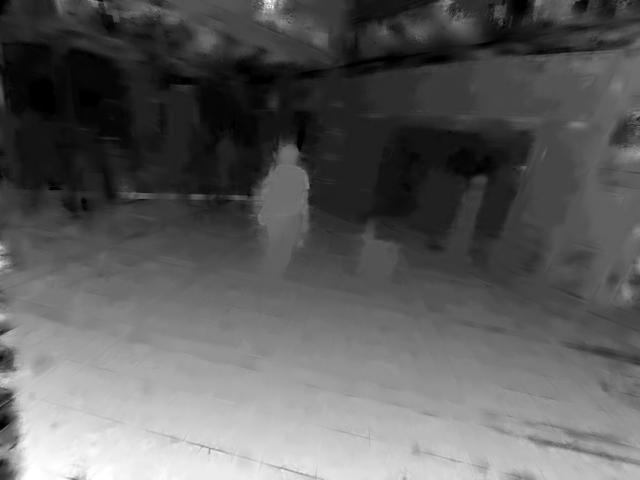} & \centering\includegraphics[width=0.9cm]{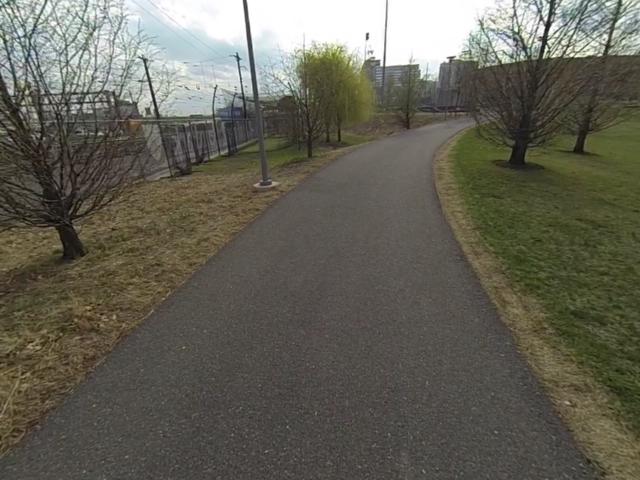}\includegraphics[width=0.9cm]{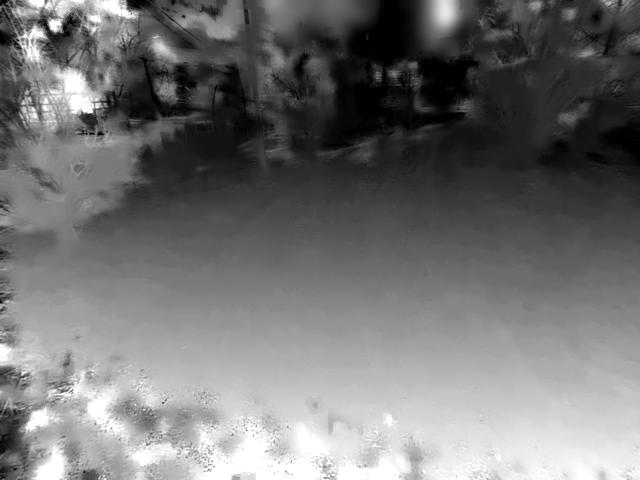} & \centering\includegraphics[width=0.9cm]{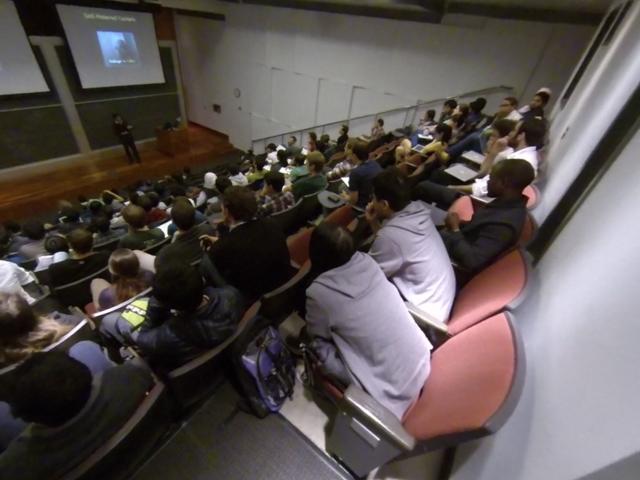}\includegraphics[width=0.9cm]{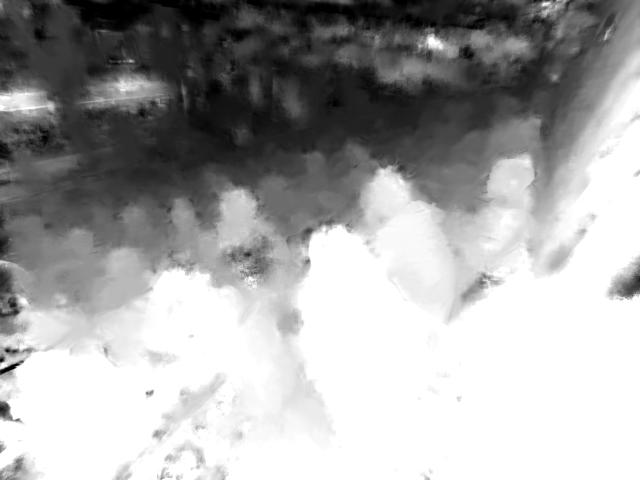} & \centering\includegraphics[width=0.9cm]{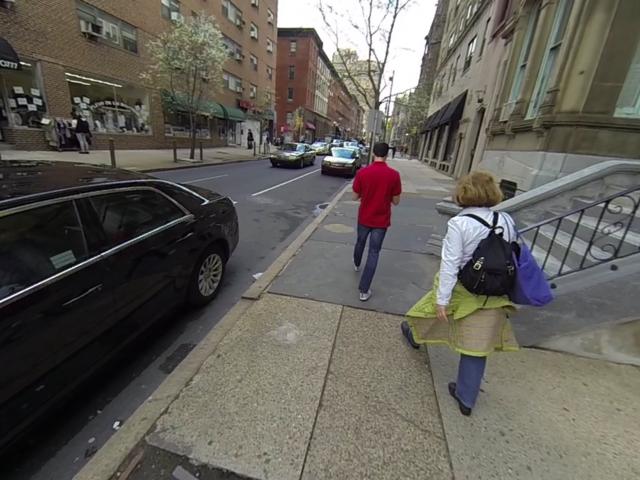}\includegraphics[width=0.9cm]{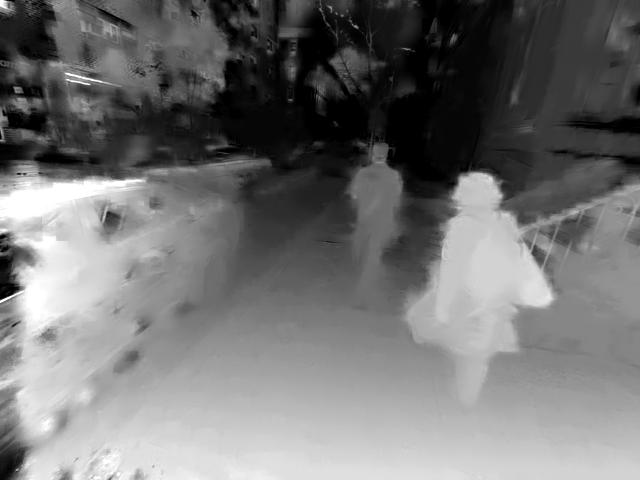} & \centering\includegraphics[width=0.9cm]{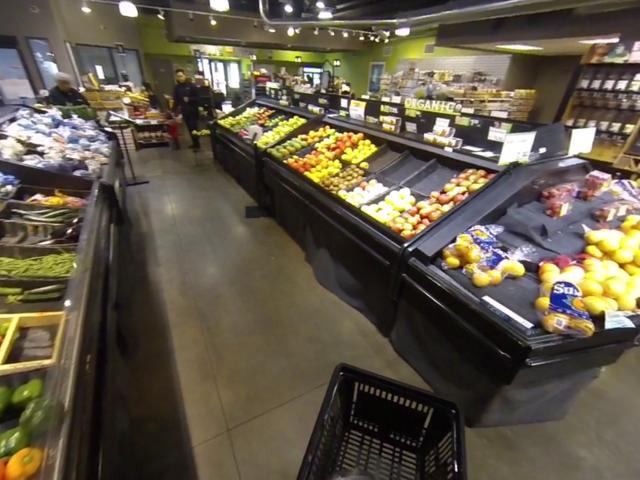}\includegraphics[width=0.9cm]{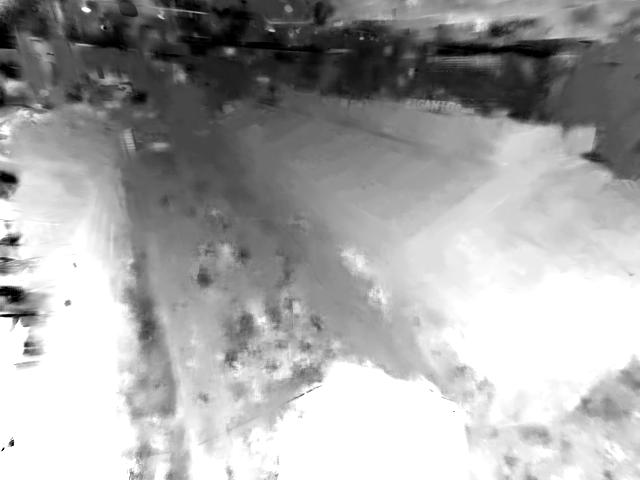} &  \centering\includegraphics[width=0.9cm]{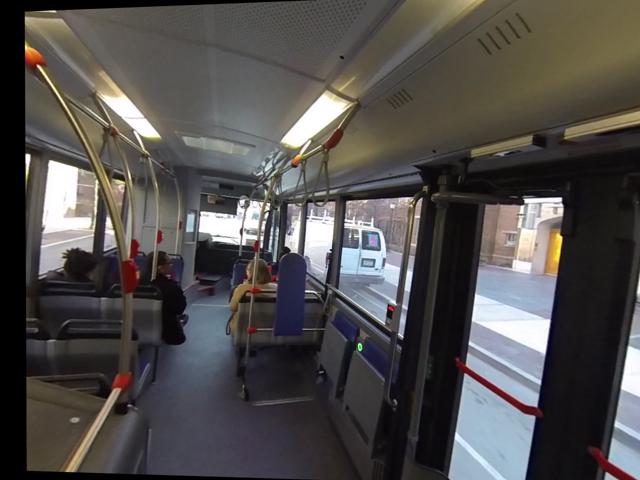}\includegraphics[width=0.9cm]{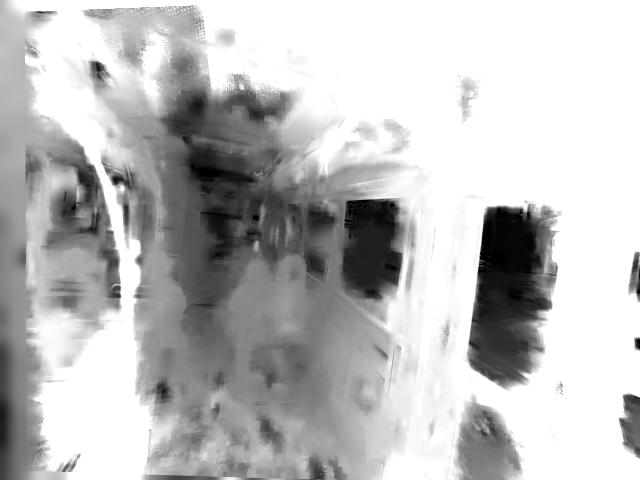} & \\
\hline
Scene & IKEA & Costco & Mall & Park & School1/2 & Downtown1/2 & Grocery1/2/3 & Bus1/2 & \\
\hline
Frames & 966 & 577 & 2683 & 3088&3754/3736 & 2856/3405 & 2858/2892/2834 & 2292/1850\\
\hline
Duration & 08:03 & 04:49 & 22:22 & 25:44 &31:17/31:08 & 23:48/28:23 & 23:49/24:06/23:37 & 19:06/15:25\\
\hline
\hline

Image Disparity & \centering\includegraphics[width=0.9cm]{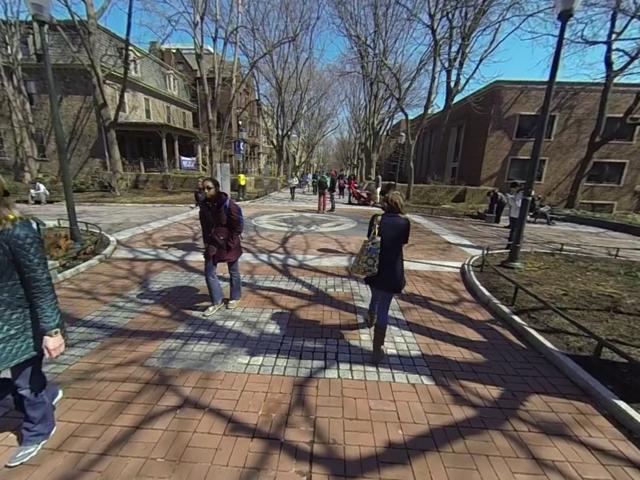}\includegraphics[width=0.9cm]{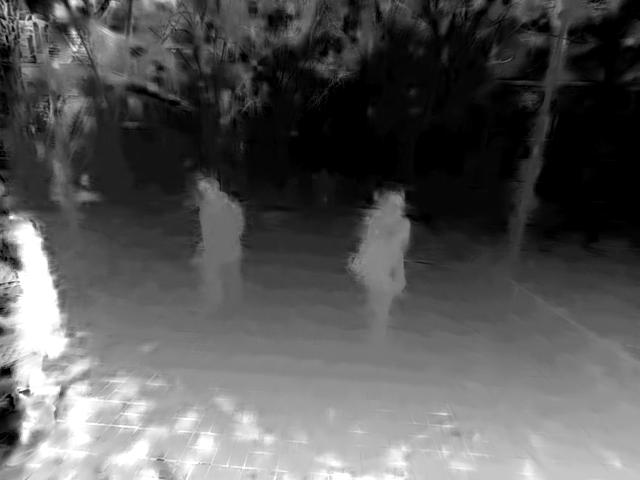} & \centering\includegraphics[width=0.9cm]{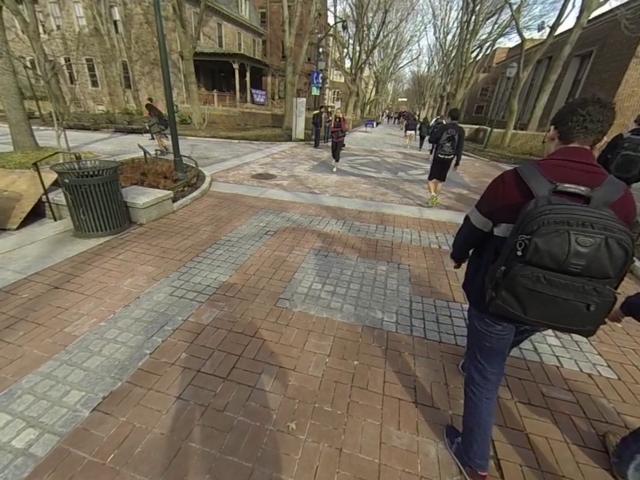}\includegraphics[width=0.9cm]{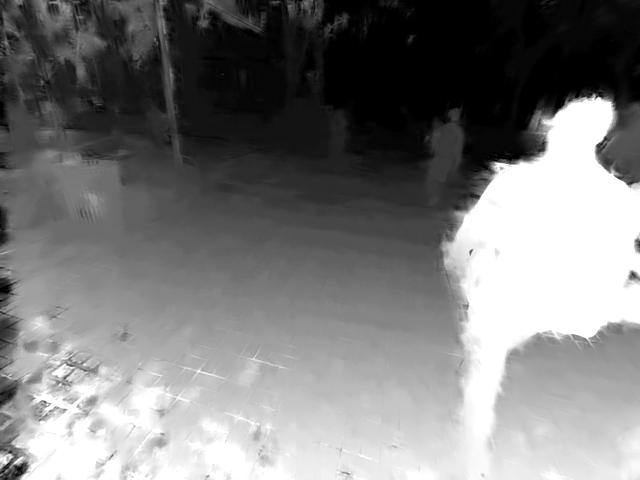} & \centering\includegraphics[width=0.9cm]{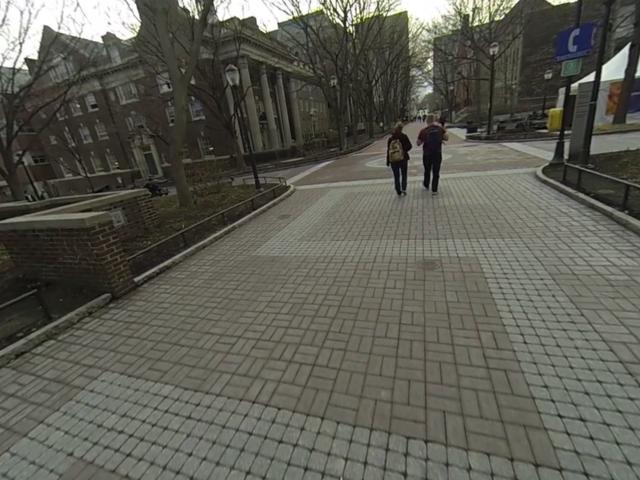}\includegraphics[width=0.9cm]{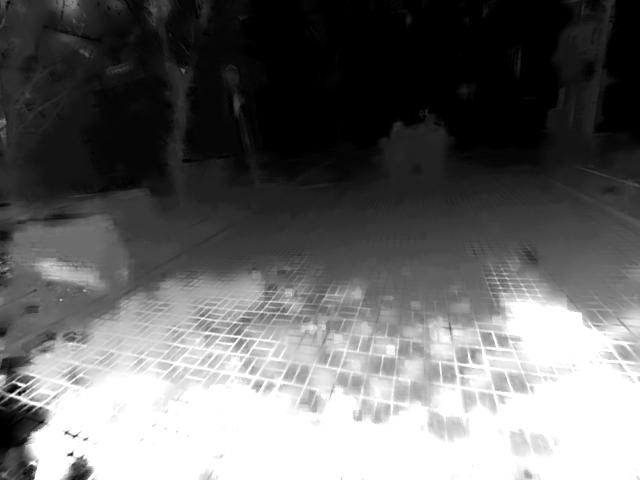} & \centering\includegraphics[width=0.9cm]{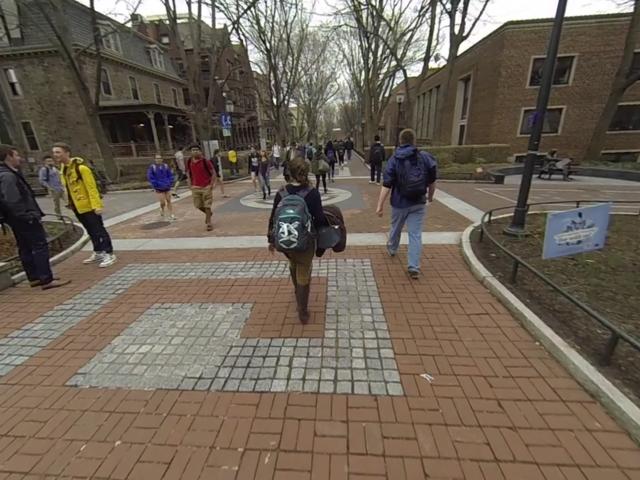}\includegraphics[width=0.9cm]{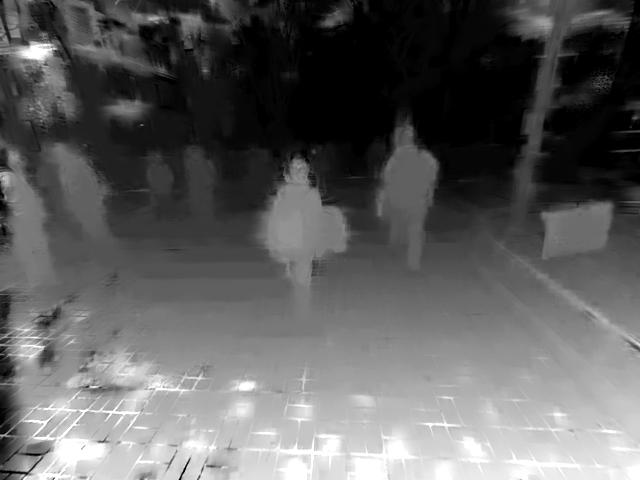} & \centering\includegraphics[width=0.9cm]{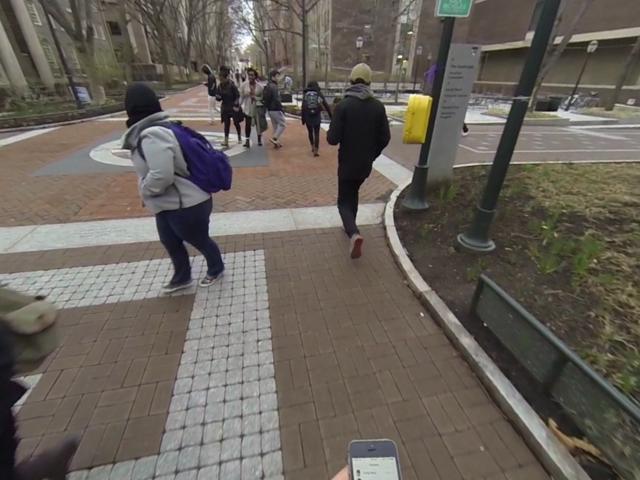}\includegraphics[width=0.9cm]{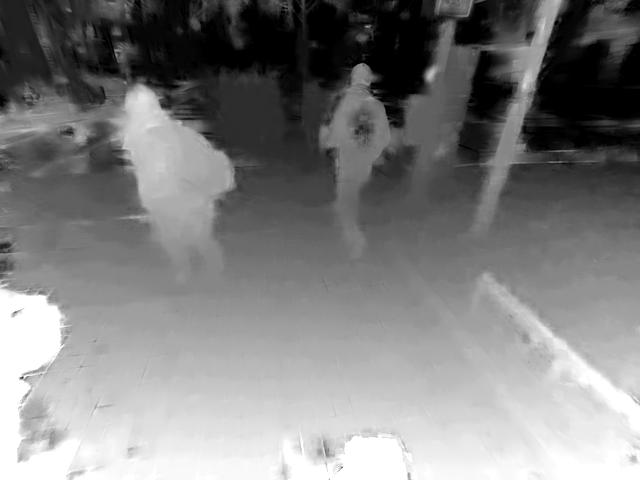} & \centering\includegraphics[width=0.9cm]{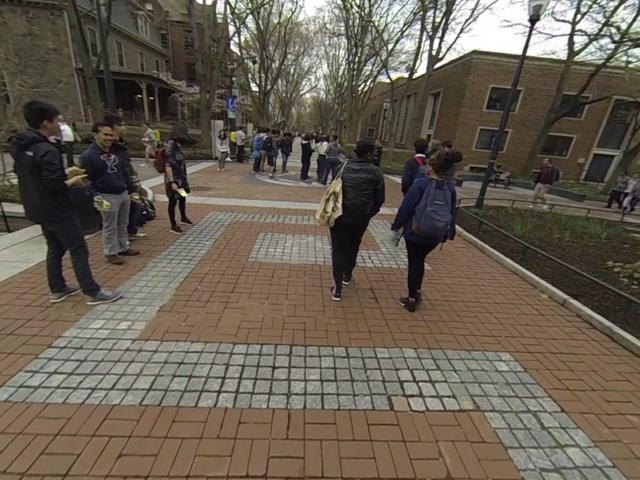}\includegraphics[width=0.9cm]{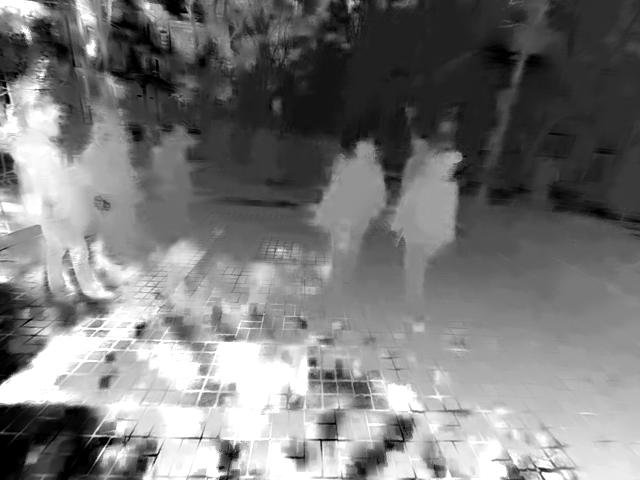} & \centering\includegraphics[width=0.9cm]{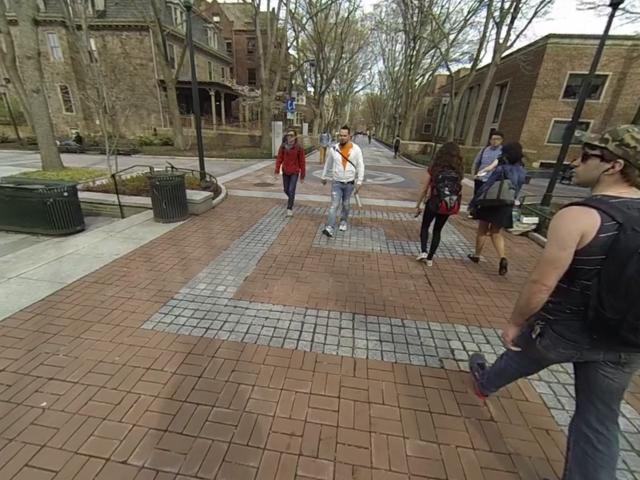}\includegraphics[width=0.9cm]{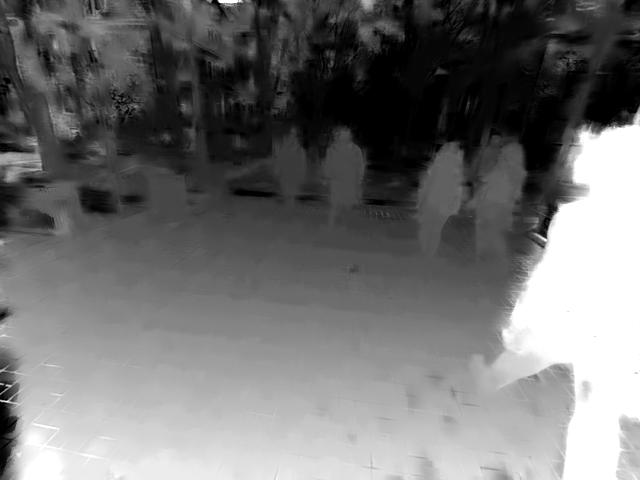} &  \centering\includegraphics[width=0.9cm]{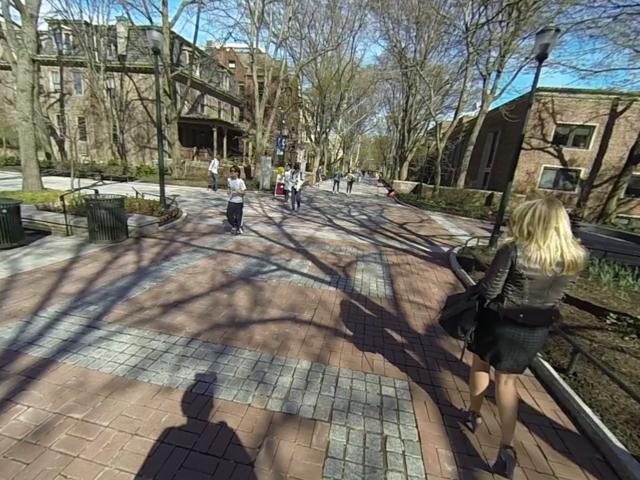}\includegraphics[width=0.9cm]{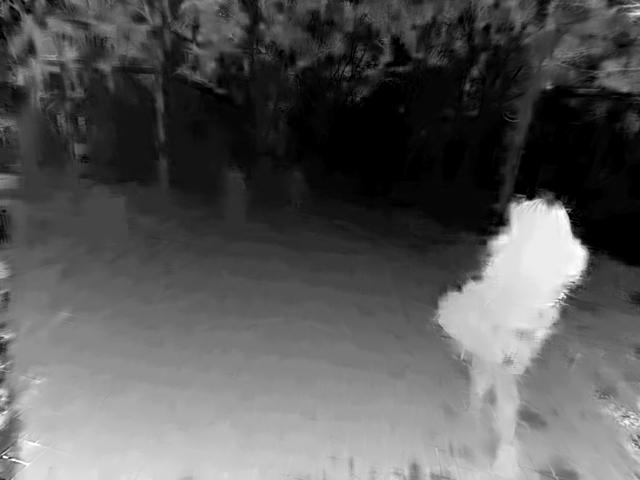} & \\
\hline
Scene    & Campus1 & Campus2 & Campus3 & Campus4 & Campus5 & Campus6 & Campus7 & Campus8 \\
\hline
Frames   & 2607    &	1884 &	1975   & 2359    &	3337   & 4034    &	2568   & 3378\\
\hline
Duration & 21:44   & 15:42   &	16:28  & 19:40   &	27:49  & 33:37   &	21:24  & 28:09\\
\hline

\end{tabular}
\caption[EgoMotion dataset]{EgoMotion dataset}\label{table:dataset}
\end{table*}

\subsection{Data Analysis} \label{Sec:analysis}

We define the EgoSpace map with respect to a gaze direction, which allows us to canonicalize all trajectories  in one coordinate system and further to represent it with compact bases. This stems from a primary conjecture: a gaze direction is aligned with ego-motion. In this section, we empirically prove the conjecture from our EgoMotion dataset.

\begin{figure}[th]
  \centering  
      \subfigure[Attention]{\label{Fig:pitch}\includegraphics[height=0.14\textheight]{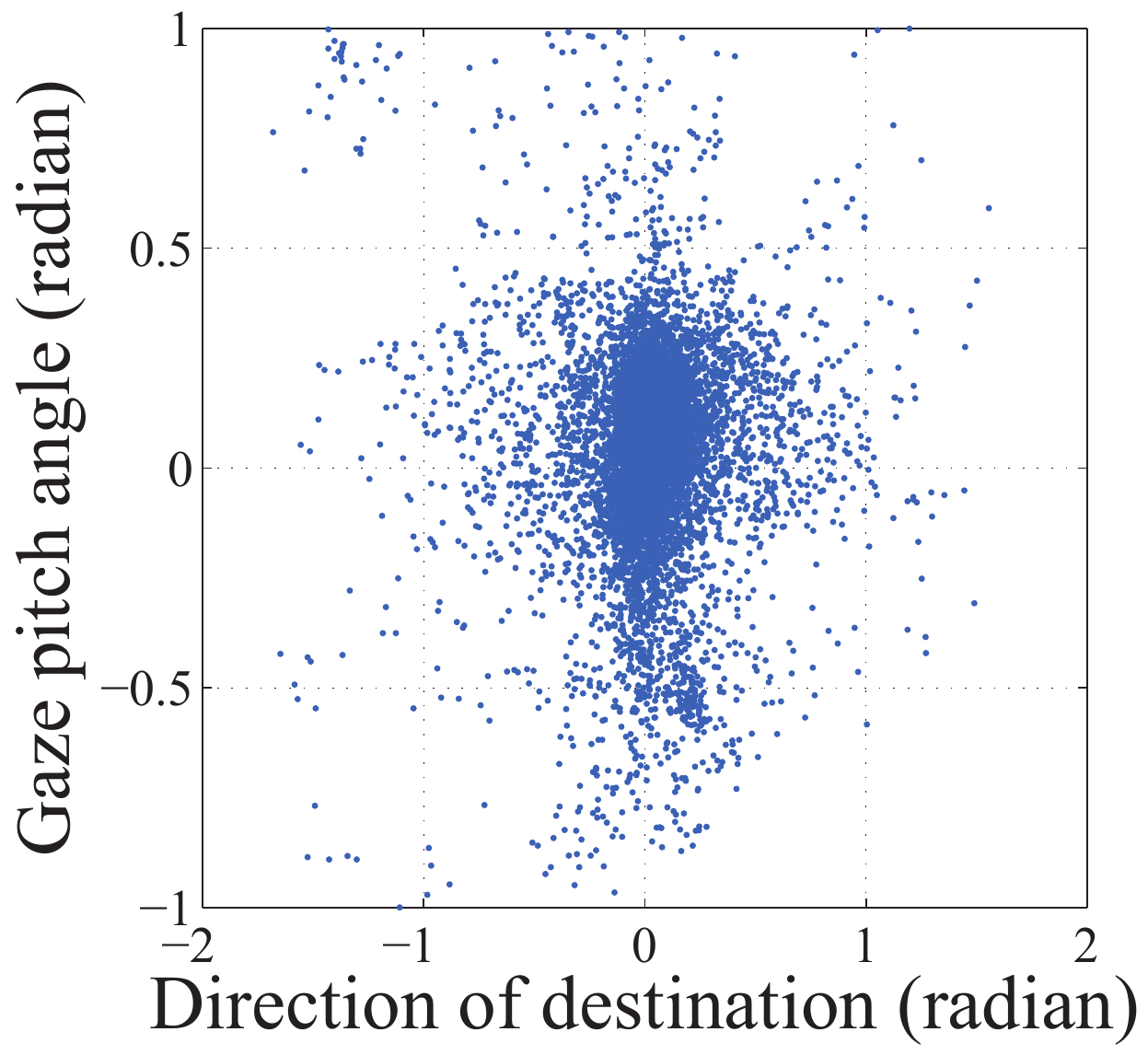}} 
      \subfigure[Yaw distribution]{\label{Fig:yaw}\includegraphics[height=0.14\textheight]{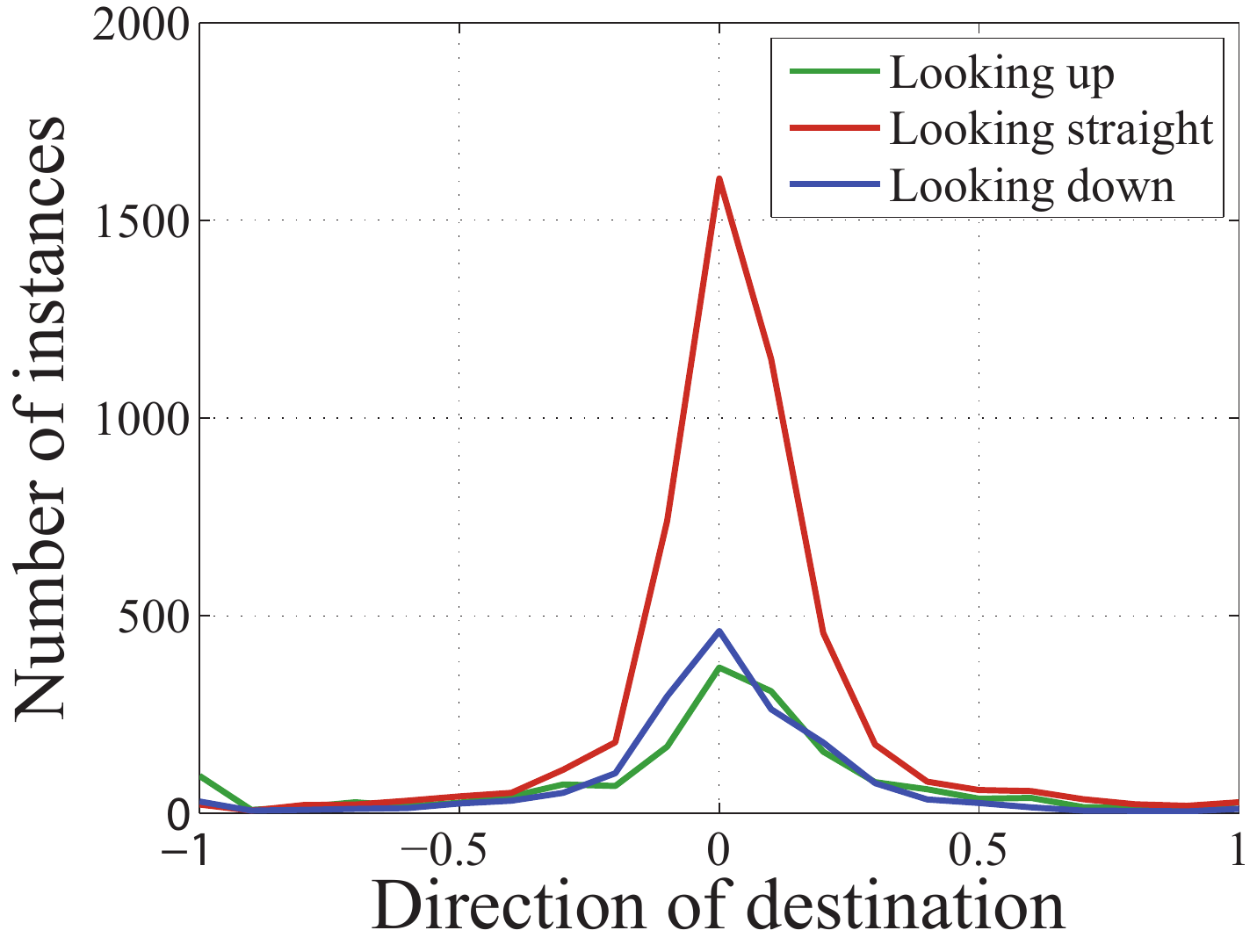}} 
  \caption{From our dataset, we empirically prove that the gaze direction is highly correlated with the direction of destination, i.e., we look where we go.} 
  \label{Fig:explanation}
\end{figure}

We compute the pitch angle of a gaze direction by calibrating the relationship between the first person camera and gaze direction~\cite{park:2012}. The pitch angle is $\cos^{-1}v_z$ by definition in Section~\ref{Sec:feaure} and the position after 10 seconds is used to measure the direction of destination. Figure~\ref{Fig:pitch} shows a distribution of the direction of destination with respect to the gaze pitch angle, which indicates that the gaze direction is aligned with the pitch axis. Figure~\ref{Fig:yaw} shows a yaw distribution of the direction of destination given pitch angle (a horizontal cross section of Figure~\ref{Fig:pitch}). This also indicates that gaze direction is highly correlated with the direction of destination. 
\vspace{-2mm}
\section{Result}\vspace{-2mm}
\begin{figure*}[t]
  \centering  
    \includegraphics[width=0.95\textwidth]{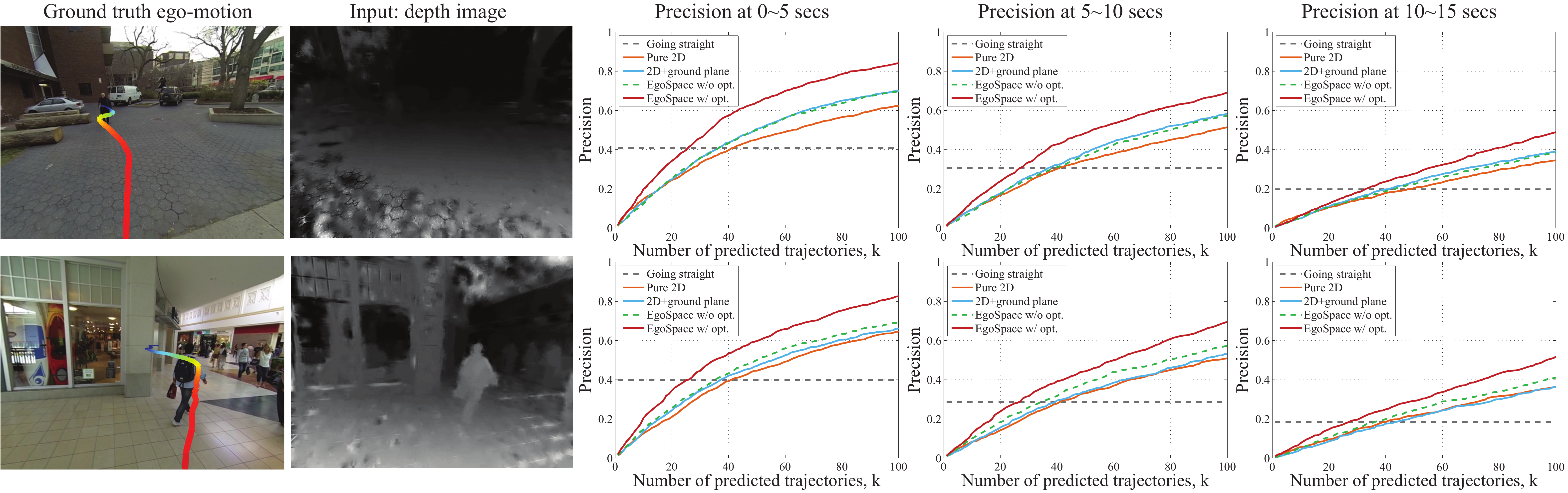}
  \caption{We compare our method with four baseline representations: (1) Going straight; (2) Pure 2D: no EgoSpace map without adaptation of the ground plane by the test scene; (3) 2D + ground plane: no EgoSpace map with adaptation of the ground plane by the test scene; (4) EgoSpace without trajectory optimization. Our method outperforms other representations.}
  \label{Fig:quant}
\end{figure*}

We apply our method to predict ego-motion and hidden space in real world scenes by leveraging the EgoMotion dataset. We divide all scenes into two categories: indoor and outdoor scenes as ego-motion has different characteristics, e.g., speed and scene layout. Note that for all evaluations, we predict a scene that is not included in training data, i.e., training and testing scenes are completely separated.

\vspace{-2mm}
\subsection{Quantitative Evaluation}\vspace{-2mm}
We quantitatively evaluate our trajectory prediction by comparing with ground truth trajectories achieved by 3D reconstruction of the first person camera. Our evaluation addresses the future localization problem.

Multiple trajectories are often equally plausible, e.g., Y-junction, while one ground truth trajectory is available per image. This results in a large prediction error. To address this multiple path configuration, we measure predictive precision---how often one of our predicted trajectories aligns with the ground truth trajectory, i.e., $prec.=\sum_{i=1}^N \mathcal{D}_i/N$, where $N$ is the number of testing images. $\mathcal{D}_i=1$ if $\min_k \max_t \|\widehat{\mathbf{X}}_t-\mathbf{X}_t^k\| < \epsilon$, and $\mathcal{D}_i=0$ otherwise where $\mathbf{X}_t^k$ is the location at the $t^{\rm th}$ time instant of the $k^{\rm th}$ predicted trajectory and $\widehat{\mathbf{X}}$ is the ground truth trajectory. We set $\epsilon = 1.5m$.
Note that unlike previous approaches measured a spatial distance between trajectories~\cite{kitani:2012}\footnote{A dynamic time warping was used to handle a time scale.}, our evaluation measures a spatiotemporal distance between trajectories because the time scale also needs to be considered.

Four baseline methods\footnote{These baseline algorithms are designed by ours because no previous algorithm exists to predict the trajectories of ego-motion} are used to compare our approach: one method solely based on gaze direction, two methods with a subsampled depth image at the same resolution of our EgoSpace map, and one method with EgoSpace map but without trajectory refinement by Equation~(\ref{Eq:min}). (1) Going straight: we generate a trajectory aligned with the gaze direction to test gaze bias; (2) Pure 2D: we retrieve a set of trajectories using KNN solely based on a subsampled depth image; (3) 2D+ground plane: we retrieve trajectories using the subsampled depth image but transform the coordinates of the trajectories such that they lie on the ground plane of the test image. This coordinate transform takes into account the 3D camera direction with respect to the ground plane of the test image; (4) EgoSpace w/o trajectory optimization: the trajectories are retrieved by the EgoSpace map but no adaptation to the test image by Equation~(\ref{Eq:min}). In fact, this method provides an initialization of our predicted trajectories.

Figure~\ref{Fig:quant} shows evaluations on indoor and outdoor depth images. We retrieve $k$ neighbors from dataset and measure precision. Our method outperforms the baseline algorithms with large margin. These experiments indicate that the EgoSpace representation has strong predictive power comparing to the camera pose oriented feature produced by the subsampled depth image. Also the scene adaptation by the trajectory optimization allows us to produce more accurate prediction (see the performance gap from the initialization). As noted in Section~\ref{Sec:analysis}, a gaze direction is a good predictor but it is not strong enough to predict a long term behavior. Note that the precision at early k may be significantly improved by using N-best algorithms~\cite{park:2011} based on homotopy class~\cite{gong:2011} because KNN retrieves many redundant trajectories. In Table~\ref{table:recon}, we measure the average precision across all scenes in Section~\ref{Sec:data}.



\begin{table}[h]
\centering
\tiny
\begin{tabular}{l|c|c|c|c|c|c|c|c|c}
\hline
\multirow{2}{*}{Indoor}&\multicolumn{3}{c|}{0$\sim$5 secs}&\multicolumn{3}{c|}{5$\sim$10 secs }&\multicolumn{3}{c}{10$\sim$15 secs}\\\cline{2-10}
&k=100&k=60&k=30&k=100&k=60&k=30&k=100&k=60&k=30\\
\hline
Going straight&\multicolumn{3}{c|}{0.571}&\multicolumn{3}{c|}{0.221}&\multicolumn{3}{c}{0.124}\\\cline{2-10}
Pure 2D&0.643&0.507&0.308&0.524&0.379&0.217&0.346&0.229&0.123\\ 
2D+Ground plane&0.710&0.556&0.367&0.561&0.413&0.267&0.384&0.261&0.162\\ 
EgoSpace w/o opt.&0.690&0.534&0.341&0.570&0.265&0.255&0.401&0.265&0.156\\
EgoSpace w/ opt.&\textbf{0.825}&\textbf{0.687}&\textbf{0.458}&\textbf{0.693}&\textbf{0.543}&\textbf{0.331}&\textbf{0.482}&\textbf{0.347}&\textbf{0.192}\\
\hline
\hline
\multirow{2}{*}{Outdoor}&\multicolumn{3}{c|}{0$\sim$5 secs}&\multicolumn{3}{c|}{5$\sim$10 secs }& \multicolumn{3}{c}{10$\sim$15 secs} \\\cline{2-10}
&k=100&k=60&k=30&k=100&k=60&k=30&k=100&k=60&k=30\\
\hline
Going straight&\multicolumn{3}{c|}{0.443}&\multicolumn{3}{c|}{0.259}&\multicolumn{3}{c}{0.103} \\\cline{2-10}
Pure 2D&0.535&0.506&0.303&0.417&0.391&0.218&0.267&0.255&0.142\\ 
2D+Ground plane&0.554&0.554&0.350&0.425&0.407&0.244&0.293&0.261&0.135\\ 
EgoSpace w/o opt.&0.567&0.527&0.329&0.432&0.399&0.233&0.289&0.250&0.141\\
EgoSpace w/ opt.&\textbf{0.683}&\textbf{0.666}&\textbf{0.441}&\textbf{0.538}&\textbf{0.522}&\textbf{0.298}&\textbf{0.373}&\textbf{0.355}&\textbf{0.171}\\\hline
\end{tabular}
\caption[Average precision (k is the number of neighbors)]{Average precision (k is the number of neighbors)}
\label{table:recon}
\end{table}

\vspace{-4mm}
\noindent\textbf{Occluded Space Discovery} We quantitatively evaluate our occluded space discovery by measuring detection rate, $D/N$ where $D$ is the number of true positive detection and $N$ the total number of detection produced by the space discovery. We threshold the likelihood of the occluded space, $\psi$, from Equation~(\ref{Eq:discovery}) and manually evaluate whether the detection is correct. Note that no ground truth label is available unless the camera wearer already had passed through the space. The detection rate in Table~\ref{table:detection} indicates that our method predicts the outdoor scenes better than the indoor scenes. This is because the indoor scenes such as Grocery and IKEA, the camera wearer had a number of close interactions with objects such as shelves or products where the view of the scenes are substantially limited.   
\vspace{-3mm}
\begin{table}[h]
\centering
\scriptsize
\begin{tabular}{c|c|c|c}
\hline
Indoor & Mall I & Grocery & IKEA\\
\hline
Detection rate & 0.5882 & 0.2371 & 0.3937\\
\hline\hline
Outdoor & Park  & Bus stop & Walk\\
\hline
Detection rate & 0.6234 & 0.6593 & 0.6338\\
\hline
\end{tabular}
\caption[Detection rate]{Detection rate}
\label{table:detection}
\end{table}

\vspace{-6mm}
\subsection{Qualitative Evaluation} \label{Sec:qual}\vspace{-2mm}
We apply our method on real world examples to predict a set of plausible trajectories of ego-motion and the occluded space by foreground objects. Our training dataset is completely separated from testing data, e.g., Grocery scene was trained to predict IKEA scene.
Given a depth image, we estimate the ground plane by a RANSAC based plane fitting with gravity and height prior. This ground plane is used to define the EgoSpace map with respect to the camera direction\footnote{The yaw angle of the gaze direction is assumed to be aligned with the camera direction.}. 

\begin{figure*}[h]
  \centering  
    \includegraphics[width=\textwidth]{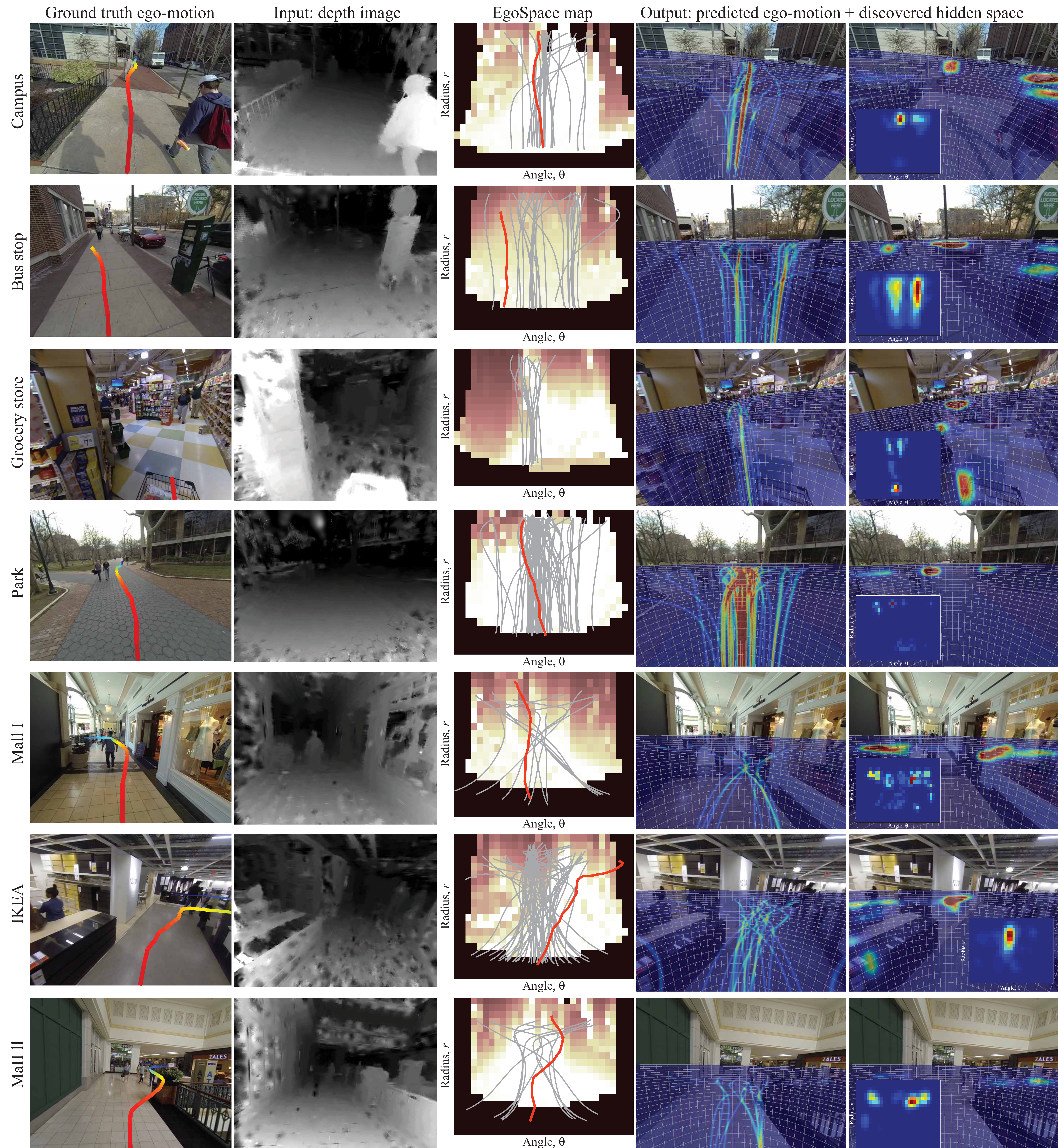}
  \caption{Given a depth image (the second column), we predict a set of plausible trajectories of ego-motion (the forth column) and discover the occluded space (the fifth column) using the EgoSpace map (the third column: predicted trajectories (gray) and ground truth trajectory (red)). The first column shows an image with ground truth trajectory of ego-motion measured by 3D reconstruction of a first person camera (time is color-coded). For more scene description, see Section~\ref{Sec:qual}.
} 
  \label{Fig:qual}
\end{figure*}

Figure~\ref{Fig:teaser} and Figure~\ref{Fig:qual} illustrate our results from the EgoMotion dataset. In a testing phase, only a depth image was used while 3D reconstruction of camera poses were used in the training phase. In Figure~\ref{Fig:qual}, we show (1) image and ground truth ego motion; (2) input depth image; (3) EgoSpace map overlaid with the predicted trajectories (gray) and ground truth trajectory (red); (4) reprojection of the trajectories; (5) reprojection of occluded space computed by the EgoSpace map (inset image). For all scenes, our method predicts the plausible trajectories that pass through unexplored space. 

\noindent\textbf{Obstacle Avoidance} Our cost function in Equation~(\ref{Eq:min}) minimizes cost difference between trajectories from training data and testing data. This precludes a trajectory passing through an object unless the retrieved trajectory was partially occluded. EgoSpace map captures the obstacle avoidance as shown in Campus and Grocery.

\noindent\textbf{Multiple Plausible Trajectories} Our prediction produces a number of plausible trajectories that conform to the testing scene. Trifurcated trajectories in Campus; bifurcated trajectories in Bus stop; and multiple directions of trajectories in Mall I.

\noindent\textbf{Occluded Space Discovery} The space occluded by foreground objects is discovered by the predicted trajectories. The space inside of the shop and behind the person in Figure~\ref{Fig:teaser}; the space occluded by the left fence and persons in Campus; the space behind the cars and the parking vending machine in Bus stop; the space behind the persons and trees in Park; the space inside the shop and around the left corner; the space behind the column; and the space occluded by the fence.


\vspace{-3mm}
\section{Discussion}\vspace{-2mm}
In this paper, we present a method to predict ego-motion and occluded space by foreground objects from a first person depth image. EgoSpace map that encodes a likelihood of occlusion is used to represent a scene around a camera wearer. We associate a trajectory with the EgoSpace map in the training phase to predict a set of plausible trajectories given a test depth image. The trajectories that are parametrized by a linear combination of compact trajectory bases are refined to conform with the test depth image. The occluded space is detected by measuring how often the predicted trajectories invade the occluded space. 

\vspace{-2mm}
\begin{figure}[h]
  \centering  
    \includegraphics[width=0.45\textwidth]{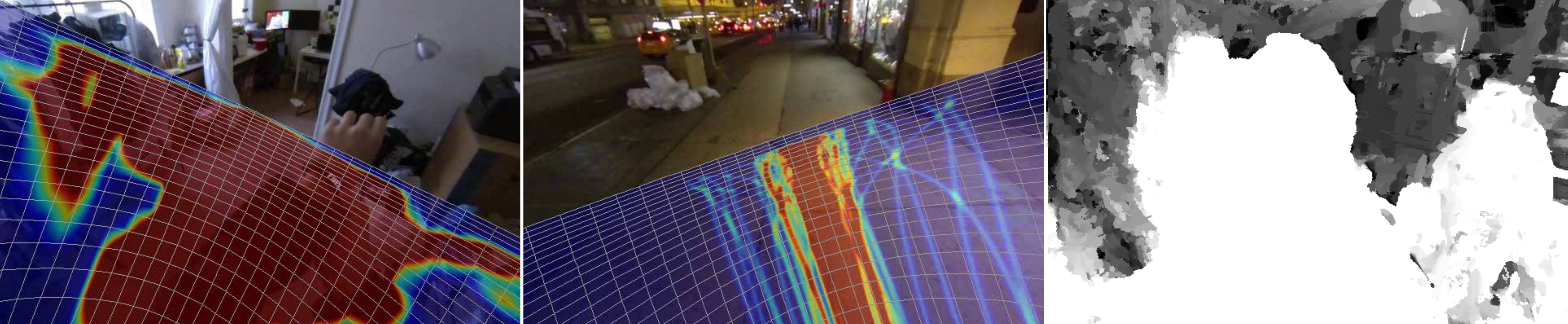}
  \caption{Our method fails due to mis-estimation ground plane, different scene distributions, and failure of depth estimation.
} 
  \label{Fig:fail}
\end{figure}
\vspace{-2mm}
\noindent\textbf{Limitation} Our framework needs three ingredients: similar scene training data, ground estimation, and depth computation. These failure cases are illustrated in Figure~\ref{Fig:fail}. 

\clearpage

{\small
\bibliographystyle{hieeetr}
\bibliography{egbib}
}

\end{document}